\documentclass{article}

\usepackage{PRIMEarxiv}

\usepackage[utf8]{inputenc} 
\usepackage[T1]{fontenc}    
\usepackage{hyperref}       
\usepackage{url}            
\usepackage{booktabs}       
\usepackage{amsfonts}       
\usepackage{nicefrac}       
\usepackage{microtype}      
\usepackage{lipsum}
\usepackage{fancyhdr}       
\usepackage{graphicx}       
\graphicspath{{media/}}     

\usepackage{multirow}
\usepackage{multicol}
\usepackage{amsmath}

\usepackage{amssymb}
\usepackage{caption}
\usepackage{subcaption}
\usepackage{amsthm}

\title{Deep Convolutional Architectures for Extrapolative Forecasts in Time-dependent Flow Problems
}

\author{
  Pratyush Bhatt, Yash Kumar \\
  Department of mechanical engineering \\
  Delhi Technological University \\
  P4X9+Q8X, Bawana Rd, Shahbad Daulatpur Village, Rohini, New Delhi, 110042, Delhi, India \\
  \texttt{bhattpratyush906@gmail.com, yashk8481@gmail.com} \\
   \And
  Azzeddine Soulaïmani \\
  Department of mechanical engineering \\
  \'Ecole de technologie sup\'erieure \\
  1100 Notre-Dame St W, Montreal, Quebec H3C1K3, Canada \\
  \texttt{Azzeddine.Soulaimani@etsmtl.ca} \\
}

\begin{document}
\maketitle

\begin{abstract}
Physical systems whose dynamics are governed by partial differential equations (PDEs) find applications in numerous fields, from engineering design to weather forecasting. The process of obtaining the solution from such PDEs may be computationally expensive for large-scale and parameterized problems. In this work, deep learning techniques developed especially for time-series forecasts, such as LSTM and TCN, or for spatial-feature extraction such as CNN, are employed to model the system dynamics for advection dominated problems. These models take as input a sequence of high-fidelity  vector solutions for consecutive time-steps obtained from the PDEs and forecast the solutions for the subsequent time-steps using auto-regression; thereby reducing the computation time and power needed to obtain such high-fidelity solutions. The models are tested on numerical benchmarks (1D Burgers' equation and Stoker’s dam break problem) to assess the long-term prediction accuracy, even outside the training domain (extrapolation). Non-intrusive reduced-order modelling techniques such as deep auto-encoder networks are utilized to compress the high-fidelity snapshots before feeding them as input to the forecasting models in order to reduce the complexity and the required computations  in the online and offline stages. Deep ensembles are employed to perform uncertainty quantification of the forecasting models, which provides information about the variance of the predictions as a result of the  epistemic uncertainties.
\end{abstract}


\section{Introduction}
\label{sec:Introduction}
Efficient numerical simulations of complex dynamical systems are needed in order to seek solutions at different times or parameter instances, especially in fluid dynamics. These systems are typically described by a set of parameterized nonlinear partial differential equations (PDEs). Obtaining numerical solutions using a high-fidelity (finite element, finite volume, or finite difference type) computational solver may be extremely expensive, as  they must create high-dimensional renderings of the solution to precisely resolve the spatial-temporal multifolds and inherent non-linearities. This method thus becomes inefficient for applications such as optimization and uncertainty quantification, where numerous simulations are required to arrive at the desired solution. Reduced-order models (ROMs) are suitable substitutions for  computationally-expensive numerical solvers, as  these methods generate a low-ranked structure of the high-dimensional snapshots, which are then utilized to model the spatio-temporal dynamics of the PDE system.  Among the various ROM techniques that have been developed, projection-based ROMs are the type employed most extensively. The method involves the generation of a reduced set of basis functions or modes such that their linear superposition effectively overlaps a low-rank approximation of the solutions. Proper Orthogonal Decomposition (POD) is the most popular method among the reduced basis class. POD utilizes  singular value decomposition (SVD) to generate an empirical basis of dominant, orthonormal modes to obtain an optimum linear subspace in which to project the system-governing PDEs \cite{rowley_dawson_2017,taira_brunton_dawson_rowley_colonius_mckeon_schmidt_gordeyev_theofilis_ukeiley_etal._2017}. Availability of the governing equations is necessary to employ intrusive ROM techniques such as the Galerkin projection \cite{berkooz_holmes_lumley_1993}, or the Petrov–Galerkin projection \cite{lozovskiy_farthing_kees_gildin_2016}, which produce an interpretable ROM defined by high-energy or dominant modes. However, scenarios where the governing equations are unavailable require the application of data-driven methods, such as non-intrusive ROM (NIROM)\cite{rezaian_biswas_duraisamy_2021, dutta_rivera-casillas_cecil_farthing_2021}. In a NIROM, the expansion coefficients for the reduced solution are obtained via interpolation on the reduced basis space spanned by the set of dominant modes. However, since the reduced dynamics generally belong to nonlinear, matrix manifolds, a variety of interpolation and regression methods have been proposed,  capable of enforcing the constraints characterizing those manifolds. Some of the methods most often employed are  dynamic mode decomposition \cite{alla_kutz_2017, rafiq_bazaz_2021, dynamic_mode_decomposition_of_a_highly_confined_shock_wave_boundary_layer_interaction_2020}, radial basis function interpolation \cite{xiao_fang_pain_navon_2017, dutta_farthing_perracchione_savant_putti_2021} and Gaussian process regression \cite{ma_yu_xiao_2022, xiao_2019}. The recent advancements in machine learning (ML) methods\cite{Karniadakis_PIML} have given rise to revolutionary approaches that effectively evaluate and expedite existing numerical models or solvers by using  online-offline computational stages. In the offline stage, the ML model updates its weights or coefficients (training) to learn the system dynamics by using the high-fidelity solutions obtained by the numerical solver, hence requiring computational power and time. In the online stage, the model uses the pre-computed/optimized weights (from the training) to obtain the solution (prediction) for a new set of input instances, and does so almost instantly with minimal computational cost. Various data-driven ML-based frameworks have been proposed to model the propagation of system dynamics in  latent space. Some of the more highly successful examples involve the use of deep neural networks (DNNs) \cite{hesthaven_ubbiali_2018}, long-short-term memory (LSTM) networks \cite{wan_vlachas_koumoutsakos_sapsis_2018, maulik_mohan_lusch_madireddy_balaprakash_livescu_2020, dutta_rivera, mut-pod} , neural ordinary differential equations (NODE) \cite{chen_rubanova_bettencourt_duvenau_2018, inproceedings_Dutta_Neural, mut-pod},  and temporal convolutional networks (TCNs) \cite{wu_sun_chang_zhang_arcucci_guo_pain_2020, xu_duraisamy_2020}. 

Significant work has been carried out recently on predicting solution instances outside the training domain for a variety of fluid problems with discontinuities, wave propagation, and advection-dominated flows. Liu et al \cite{liu_fu_xiao_stefanescu_sharma_zhu_sun_wang_2022} presented a predictive data assimilation framework based on the Ensemble Kalman Filter (EnKF) and the DDROM model, which uses an autoencoder network for the compression of high-dimensional dynamics to lower dimensional space and then the LSTM method to model the fluid dynamics in latent space. The model capabilities were estimated using 2D Burgers’ equation and flow past a cylinder test case. Maulik et al. \cite{maulik_lusch_balaprakash_2021} proposed a Convolutional Autoencoder (CAE) for compression and a recurrent LSTM network for time evolution on the reduced space. The CAE-LSTM model was capable of reconstructing the sharp profile of the advecting Burgers' equation more accurately than the POD-Galerkin technique. Dutta et al. \cite{dutta_rivera} utilized an  advection-aware (AA) autoencoder network that learns nonlinear embeddings of the high-fidelity system snapshots using an arbitrary snapshot from the dataset, and then models the latent space dynamics using an LSTM network to make predictions for the linear advection and Burgers' problem. Cheng et al. \cite{cheng_xu_feng_2022} used the POD-ANN model, in which they performed a priori dimension reduction on the high-fidelity dataset and parameterization with  an artificial neural (ANN) network to solve the strongly non-linear Allen-Cahn equations and the cylinder flow problem. Heaney et al.\cite{heaney_wolffs} proposed an AI-DDNIROM framework, capable of making predictions for spatial domains, significantly larger than the training domain, using a domain decomposition approach, an autoencoder network for low-rank representation, and an adversarial network for making the predictions for flow past a cylinder and slug flow problems. Fatone et al.  \cite{mut-pod} introduced a µt-POD-LSTM ROM framework that is capable of extrapolation for time windows around 15\% those of the training domain on unsteady advection-diffusion and unsteady Navier-Stokes equation for new parameter instances. Xu et al. \cite{xu_duraisamy_2020} proposed a multi-level framework comprising a convolution autoencoder (CAE), a temporal CAE (TCAE) and a multilayer perceptron (MLP), for the purpose of parameterization, and a TCN network for auto-regressive future state predictions, and evaluated the results on problems such as Sod's-shock tube and transient ship airwakes. Wu et al. \cite{wu_sun_chang_zhang_arcucci_guo_pain_2020}  developed a POD and TCN-based neural network for  making predictions on the viscous periodic flow past a cylinder case. Abdedou et.al \cite{abdedou_stochastic} proposed two CAE architectures to compress the high-dimensional snapshot matrices obtained from numerical solvers for the Burgers', Stoker's, and dam-break equations in space and time, and performed parameterization on the compressed latent space. Jacquier et al. \cite{jacquier_abdedou_delmas_soulaimani_2021} employed uncertainty quantification methods - Deep Ensembles and Variational Inference-based Bayesian Neural Networks on the POD-ANN order-reduction method to perform predictions within and outside of the training domain on problems such as shallow water equations for flood prediction, and generated probabilistic flooding maps aware of model uncertainty. Geneva et al. \cite{geneva_zabaras_2020} presented a physics-constrained Bayesian auto-regressive CAE network that models non-linear dynamical systems (Kuramoto-Sivashinsky equation, 1D Burgers', 2D Burgers') devoid of training data, using only the initial conditions. This  reduces the computation cost tremendously and provides uncertainty quantification at each time-step.

The caveat that remains is long-term temporal extrapolation for fluid problems marked by sharp gradients and discontinuities. Our study explores forecasting convolutional architectures (LSTM, TCN, and CNN) to obtain accurate solutions for time-steps distant from the training domain, on advection-dominated test cases. The high-dimensional input snapshots matrix is first compressed in space to obtain the reduced latent vectors before they are passed  as a sequence to the forecasting models. Two types of architectures are first evaluated for space compression - MLP autoencoder and CAE autoencoder,  to identify the one that is more accurate in terms of the reconstruction  and preservation of the input information. A simple convolutional architecture is then proposed and shown to provide accurate results for the forecasts. To evaluate the epistemic uncertainties in the solutions, the methodology of deep ensembles is adopted. 

The subsequent sections of the paper are organized as follows. Section 2 describes the dataset structure along with the training and testing strategies, followed by a presentation of the autoencoders  for space compression and the forecasting convolutional architectures. In Section 3, the models are tested on two numerical cases - one-dimensional Burgers' and  Stoker's equations, which are representative of advection-dominated flows. Finally, section 4 presents a summary of the results obtained by the models, and some concluding remarks. 

\section{Methodology}
\label{sec:Methodology}

\subsection{Dataset}
\label{subsec:Dataset}

\begin{figure}[t!]
    \centering
    \includegraphics[width = \textwidth]{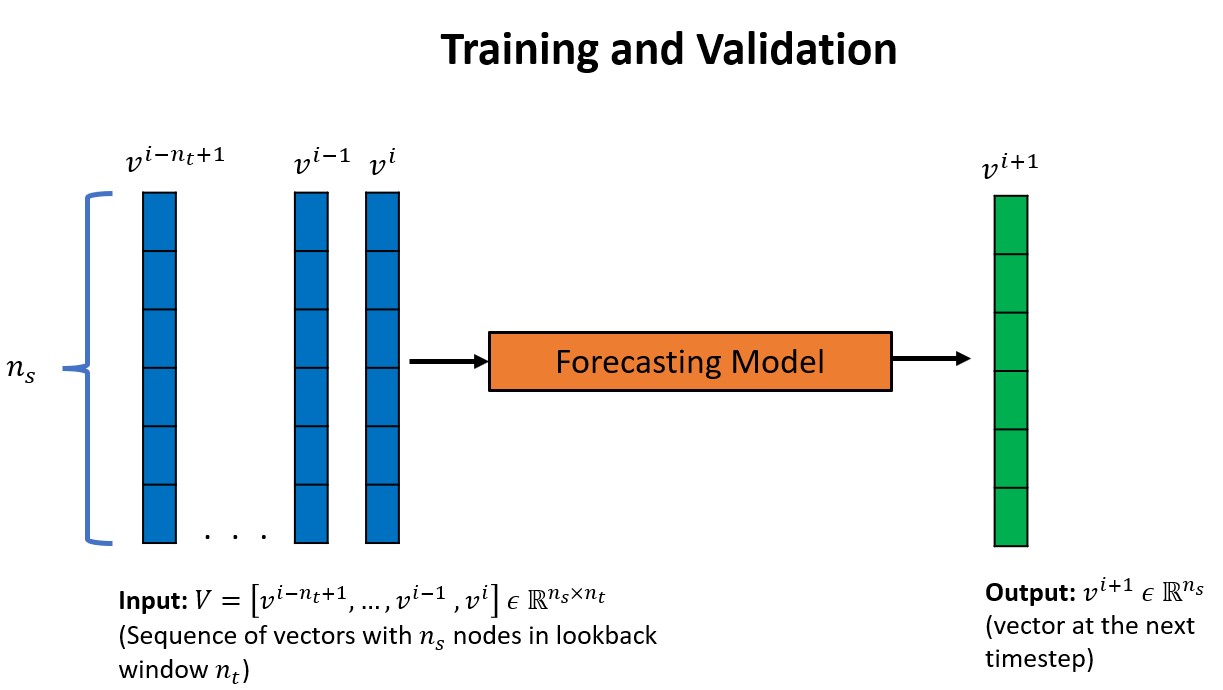}
    \caption{Training and validation method}
    \label{fig:1}
\end{figure}

\begin{figure}[t!]
    \centering
    \includegraphics[width = \textwidth]{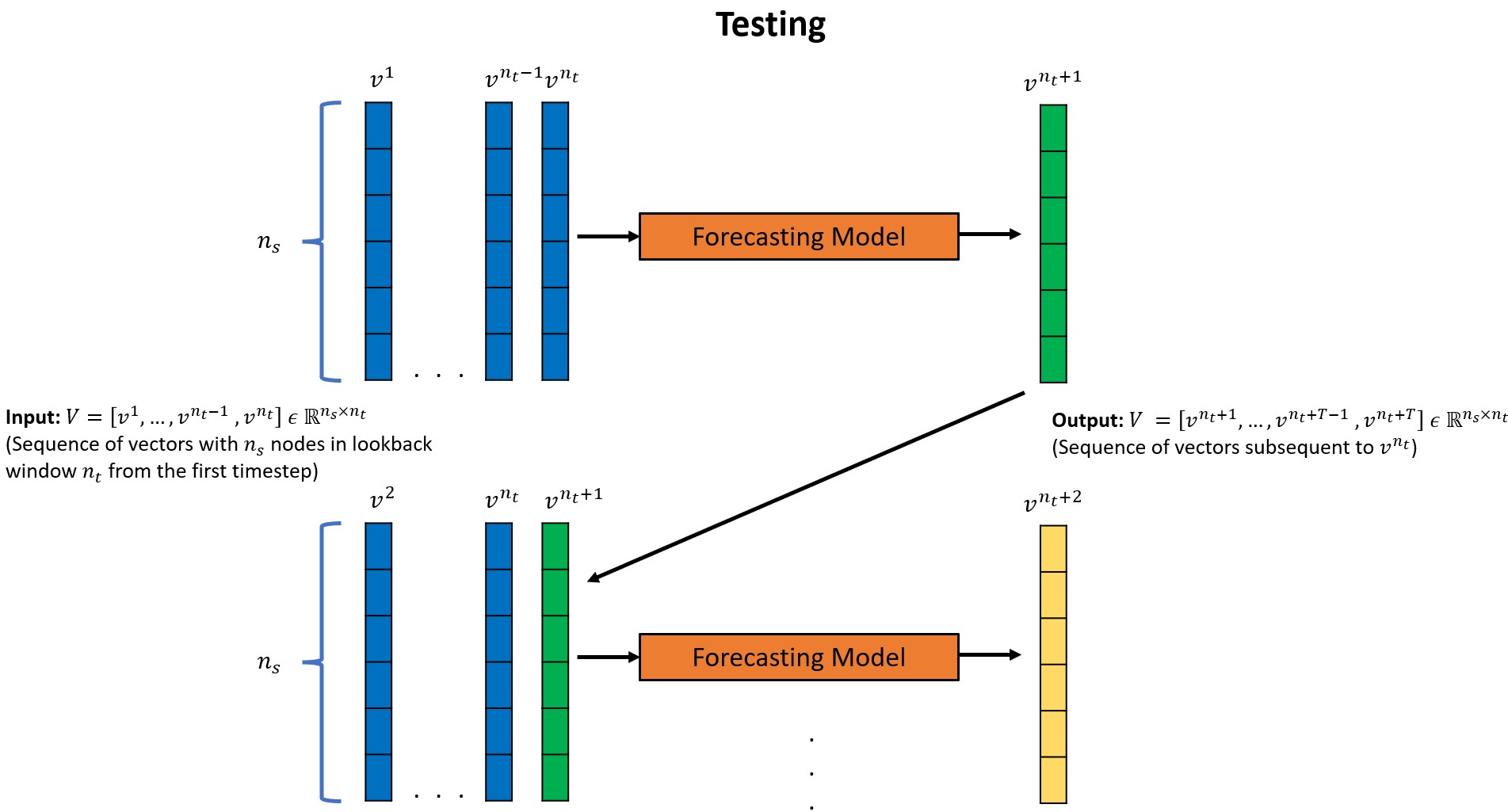}
    \caption{Autoregressive testing method for forecasting models}
    \label{fig:2}
\end{figure}

The dataset is comprised of $T$ solution vectors/snapshots: $v^{i}$ with $n_s$ nodes ($v^i \, \epsilon \, \mathbb{R}^{n_s}$) at time-steps $i$ $\epsilon$ $\{1, 2, …, T\}$ obtained using a high-fidelity PDE solver. For the autoencoder models, the output is the reconstruction of the input, therefore the training and validation input and output data are snapshot vectors $v^{i}$. For the forecasting models (Figure \ref{fig:1}), N samples are used for training; in each sample, the input is a sequence of $n_{t}$ snapshots (lookback window = $n_t$): $V = [v^{i-n_{t}+1}, …, v^{i-1}, v^{i}]$, with $V \,  \epsilon \, \mathbb{R}^{n_s \times n_t}$, and the corresponding output is the vector at the time-step immediately after the sequence end - $v^{i+1}\,  \epsilon \, \mathbb{R}^{n_s}$. 
For extrapolative testing (Figure \ref{fig:2}), a sequence of $n_{t}$ vectors from the start of the dataset, $V = [v^{1}, …,v^{n_t-1}, v^{n_{t}}]\,  \epsilon \, \mathbb{R}^{n_s \times n_t}$ is fed to the model to produce the vectors at all the subsequent time-steps: $[v^{n_{t}+1}, v^{n_{t}+2}, …, v^{T}]\,  \epsilon \, \mathbb{R}^{n_s \times (T-n_t)}$ in an auto-regressive manner, i.e, first only a single subsequent snapshot $v^{n_{t}+1}$ is predicted, which is then concatenated with previous $n_t-1$ vectors and passed to the forecasting model to produce vector $v^{n_{t}+2}$. This process is repeated in accordance with the desired number of subsequent solution vectors. 

\subsection{Non-intrusive reduced-order modelling}
\label{subsec:NIROM}
Non-intrusive ROMs (NIROMs) bypass the governing equations and utilize the full-order model solutions to develop a data-driven model, which compresses the full order data (snapshot) into a reduced-order (latent) space. The method most widely adopted to perform this utilizes deep neural network architectures called autoencoders \cite{Goodfellow-et-al-2016}. 

\begin{figure}[t!]
    \centering
    \includegraphics[width = \textwidth]{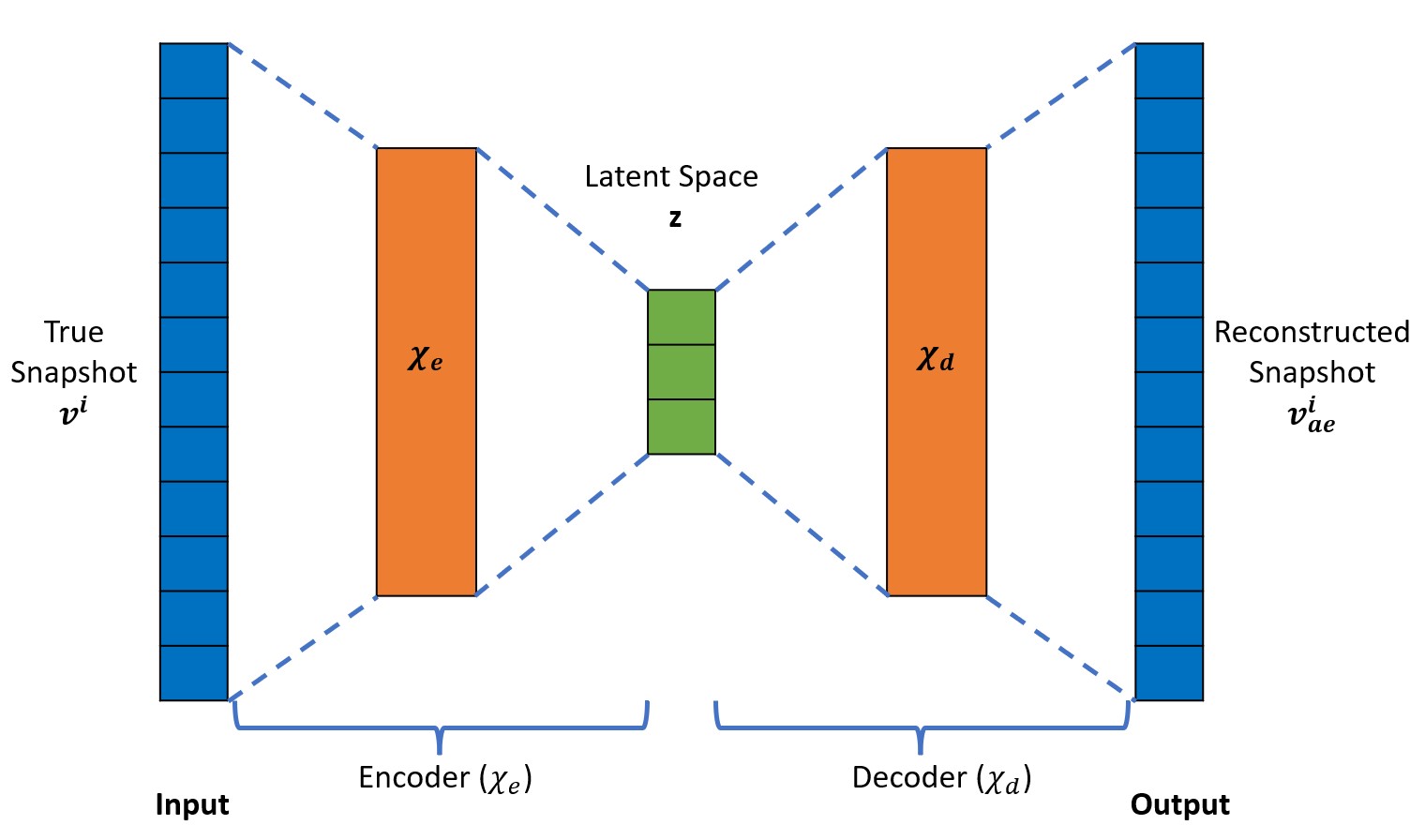}
    \caption{Autoencoder architecture}
    \label{fig:3}
\end{figure}
\begin{figure}[t!]
    \centering
    \includegraphics[width = \textwidth]{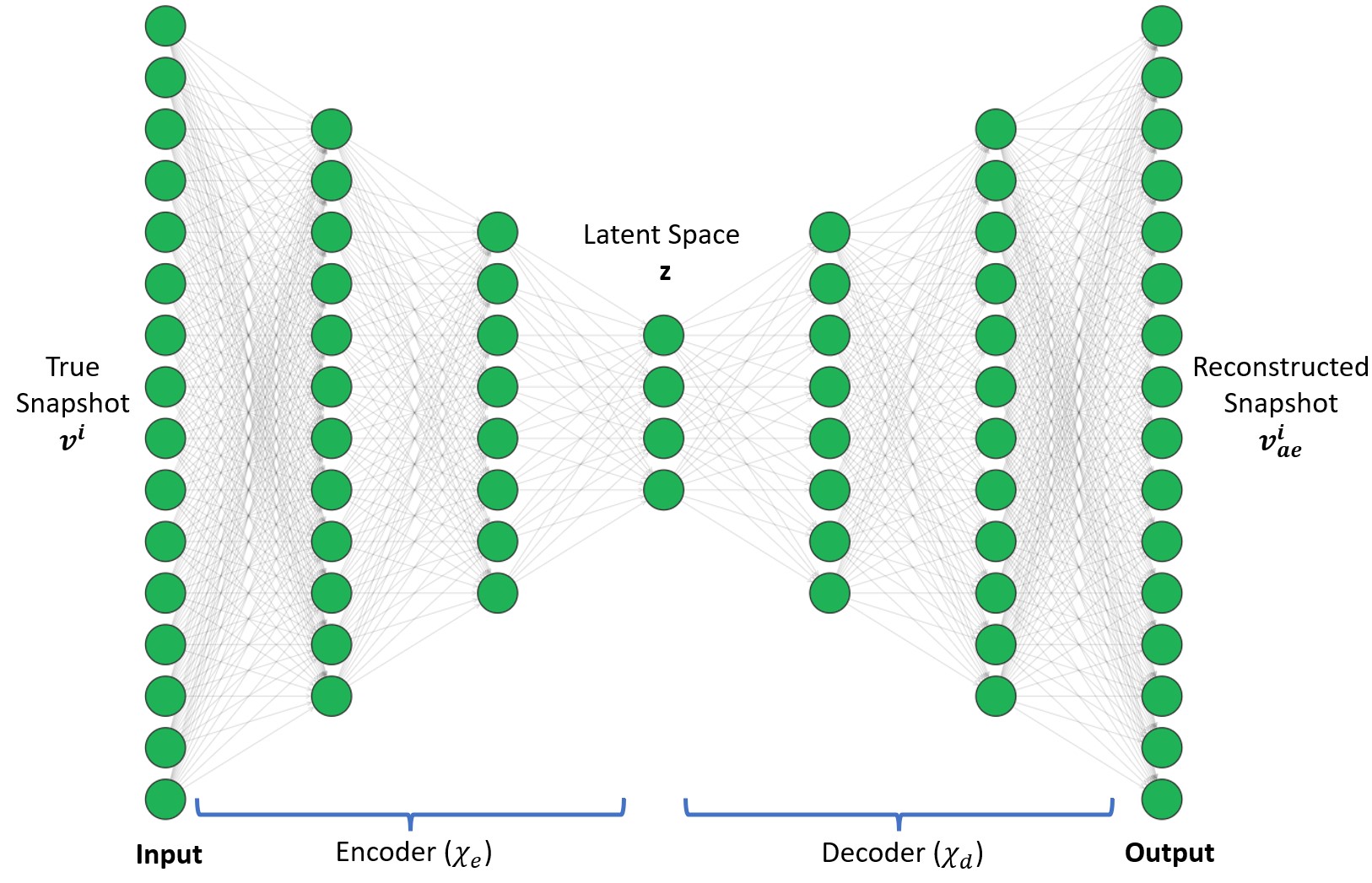}
    \caption{MLP Autoencoder architecture}
    \label{fig:4}
\end{figure}
\begin{figure}[t!]
    \centering
    \includegraphics[width = \textwidth]{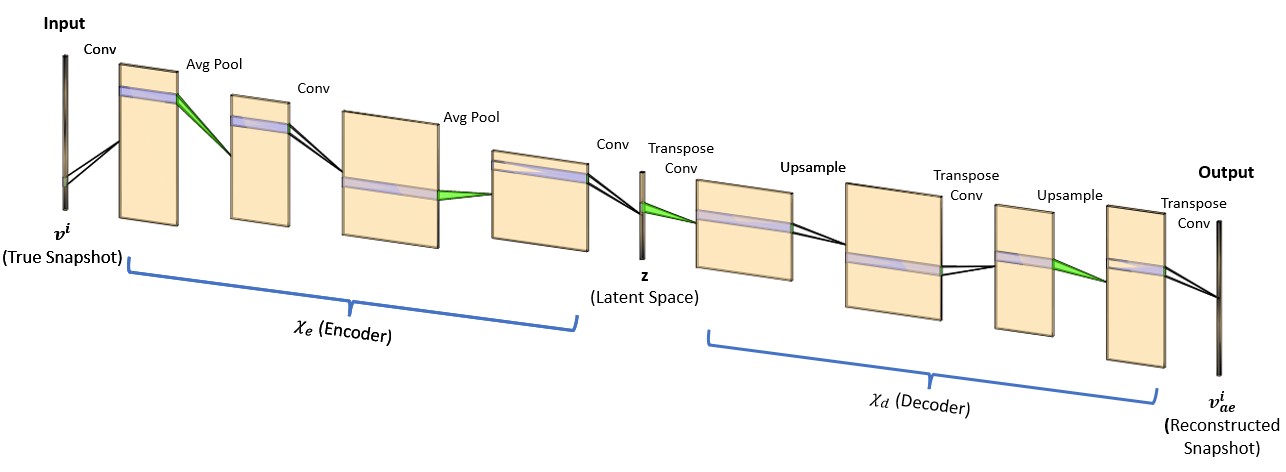}
    \caption{Convolutional Autoencoder architecture}
    \label{fig:5}
\end{figure}

\noindent An autoencoder learns the approximation of the identity mapping, $\chi$: $v^i$ → $v_{ae}^i$ such that $v^i \approx v_{ae}^i$ and $\chi$ : $\mathbb{R}^{n_s}$ → $\mathbb{R}^{n_s}$, where ${n_s}$ is the number of nodes in the solution vector $v^i$. This process is accomplished using a two-part architecture. The first part of the autoencoder network is the encoder $\chi_e$, which maps a high-dimensional input vector $v^i$ to a low-dimensional latent vector $z^i$: $z^i = \chi_e (v^i;  \theta_e )$ and $z^i$ $\epsilon$ $\mathbb{R}^m$ $(m \ll n)$. The second part is called a decoder, $\chi_d$, which maps the latent vector $z^i$ to an approximation $v_{ae}^i$ of the high-dimensional input vector $v^i$: $v_{ae}^i$ = $\chi_d (z^i; \theta_d$). The combination of these two parts yields an autoencoder network (Figure \ref{fig:3}) of the form $\chi$ : $v^i$ → $\chi_d$ ◦ $\chi_e(v^i)$. The autoencoder model is trained by computing optimal values of the parameters ($\theta_e$, $\theta_d$) that minimize the reconstruction error over all the training data \cite{dutta_rivera}:

\begin{equation}
    \theta_e, \theta_d = argmin \mathcal{L}(v^i, v_{ae}^i)
\end{equation}
\noindent where  $\mathcal{L}(v^i, v_{ae}^i)$ is a chosen measure of discrepancy between $v^i$ and its approximation $v_{ae}^i$. The restriction (dim($z^i$) = m) $\ll$ (n = dim($v^i$)) forces the autoencoder model to learn the salient features of the input data via compression into a low-dimensional space and to then reconstruct the input, instead of directly learning the identity function. Autoencoder architectures are  generally comprised of MLPs (called AAs) \cite{dutta_rivera}, convolutional neural network autoencoders (called CAEs) \cite{maulik_lusch_balaprakash_2021, xu_duraisamy_2020, abdedou_stochastic}, or a combination of both. While small-sized problems can be effectively modelled via an MLP architecture, problems involving data of high spatial complexity require CAE autoencoders for effective and accelerated spatial compression. The architecture of an MLP autoencoder,  with two fully connected dense layers (hidden layers) in the encoder network and a mirrored decoder network, is shown in (Figure \ref{fig:4}). The Convolution autoencoder consists of two convolution layers, each followed by batch normalization, swish activation, and an average pooling layer, as described in (Figure \ref{fig:5}).

\subsection{Forecasting Techniques}

The dataset (Section \ref{subsec:Dataset}) post compression by the encoder ($\chi_e$) produces $N$ samples of the form: {$Z=[z^{i-n_{t}+1}, …, z^{i-1}, z^{i}] \,  \epsilon \, \mathbb{R}^{m \times n_t}, \, z^{i+1} \epsilon \, \mathbb{R}^m$}, which are used to train the following forecasting models.

\subsubsection{Long Short-Term Memory (LSTM)}
\label{subsubsec:LSTM}
LSTM \cite{Hochreiter_LSTM} is a special type of recurrent neural network (RNN) that is well-suited for performing regression tasks based on time series data. The main difference between the traditional RNN and the LSTM architecture is the capability of an LSTM memory cell to retain information over time and an internal gating mechanism that regulates the flow of information in and out of the memory cell \cite{theodoridis_2020}. The LSTM cell consists of three parts, also known as gates, that have specific functions. The first part, called the forget gate, chooses whether the information  from the previous step in the sequence is to be remembered or can be forgotten. The second part, called the input gate, tries to learn new information from the current input to this cell. The third and final part, called the output gate, passes the updated information from the current step to the next step in the sequence. The basic LSTM equations for an input vector $v^i$ are:

\begin{figure}[t!]
    \centering
    \includegraphics[width = \textwidth]{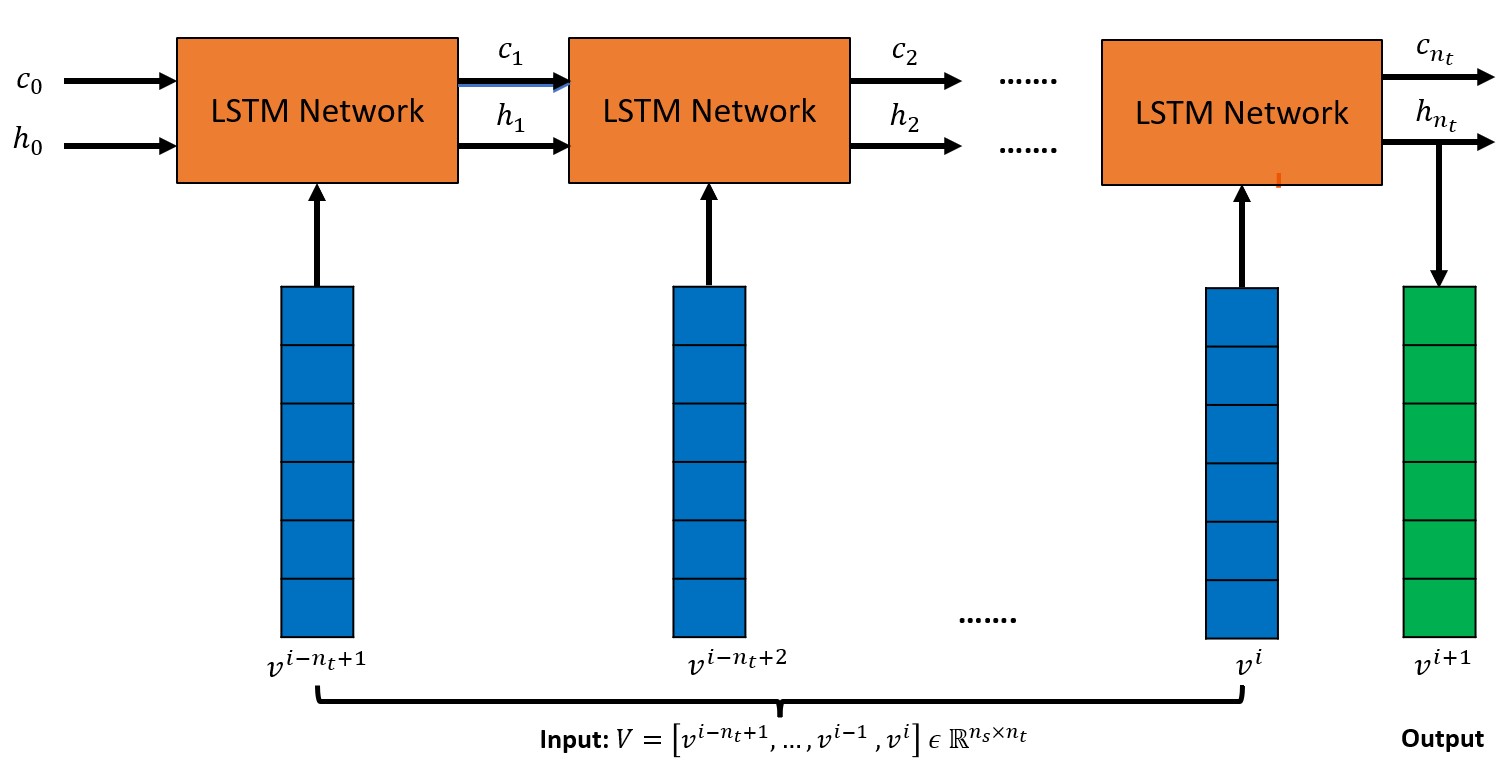}
    \caption{LSTM recurrence on vector sequence}
    \label{fig:6}
\end{figure}

\begin{equation}
    input\,gate: \zeta_{in} = \alpha_s \circ F_{in}(v^i)
\end{equation}
\begin{equation}
    forget\,gate: \zeta_{for} = \alpha_s \circ F_{for}(v^i)
\end{equation}
\begin{equation}
    cell\,state: c_i = \zeta_{for} \odot c_{i-1} + \zeta_{in}\odot (\alpha_t \circ F_a(v^i))
\end{equation}
\begin{equation}
    output\,gate: \zeta_{out} = \alpha_s \circ F_{out}(v^i)
\end{equation}
\begin{equation}
    output: h_i = \zeta_{out}\circ \alpha_t(c_i)
\end{equation}

\noindent Here, $F$ refers to a linear transformation defined by a matrix multiplication and bias addition, that is, $ F(v^i) = W v^i + b$, where W $\epsilon$ $\mathbb{R}^{h \times n_s}$ is a matrix of layer weights ($h$ is number of neurons in the LSTM cell), b $\epsilon$ $\mathbb{R}^{h}$ is a vector of bias values, and $v^i$ $\epsilon$ $\mathbb{R}^{n_s}$ is the input vector to the LSTM Cell. Also, $\alpha_s$ and $\alpha_t$ denote sigmoid and hyperbolic tangent activation functions, respectively, which are standard choices in an LSTM network, and $x \odot y$ denotes a Hadamard product of two vectors $x$ and $y$. The sequence of snapshot vectors of $n_{t}$ time-steps: $V = [v^{i-n_{t}+1}, …, v^{i-1}, v^{i}]$, with $V \,  \epsilon \, \mathbb{R}^{n_s \times n_t}$ trains the LSTM network, with recurrence over time (Figure \ref{fig:6}), to predict the subsequent vector $v^{i+1}$. The core concept of an LSTM network is the cell state $c_i$, which behaves as the “memory” of the network. It can either allow greater preservation of past information, reducing the issues of short-term memory, or it can suppress the influence of the past, depending on the actions of the various gates during the training process.

\subsubsection{Temporal Convolution Network (TCN)}
\label{subsubsec:TCN}

\begin{figure}[t!]
    \centering
    \includegraphics[width = \textwidth]{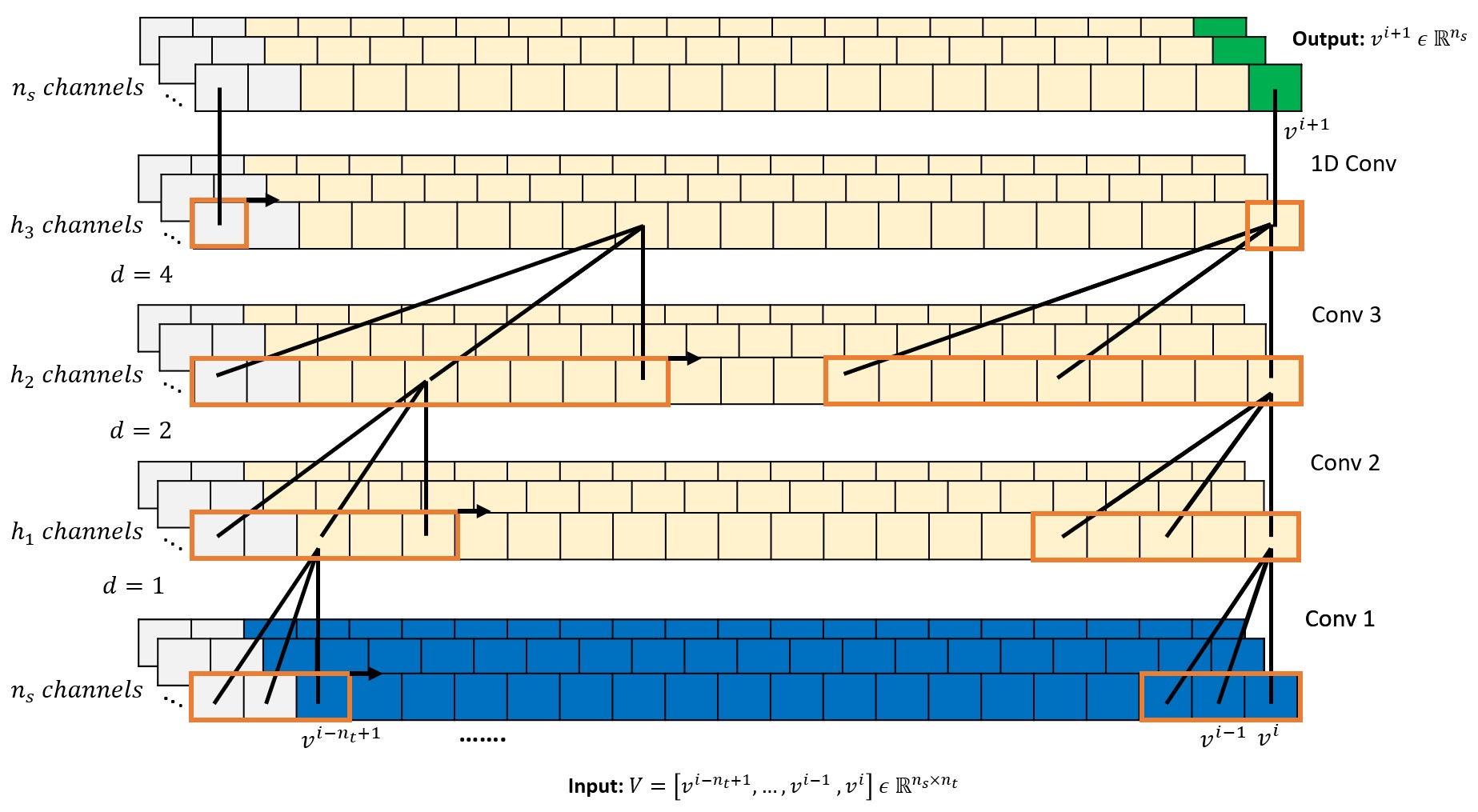}
    \caption{1D dilated filters convolving on the temporal dimension of vectors}
    \label{fig:7}
\end{figure}
\
\begin{figure}[t!]
    \centering
    \includegraphics[width = \textwidth]{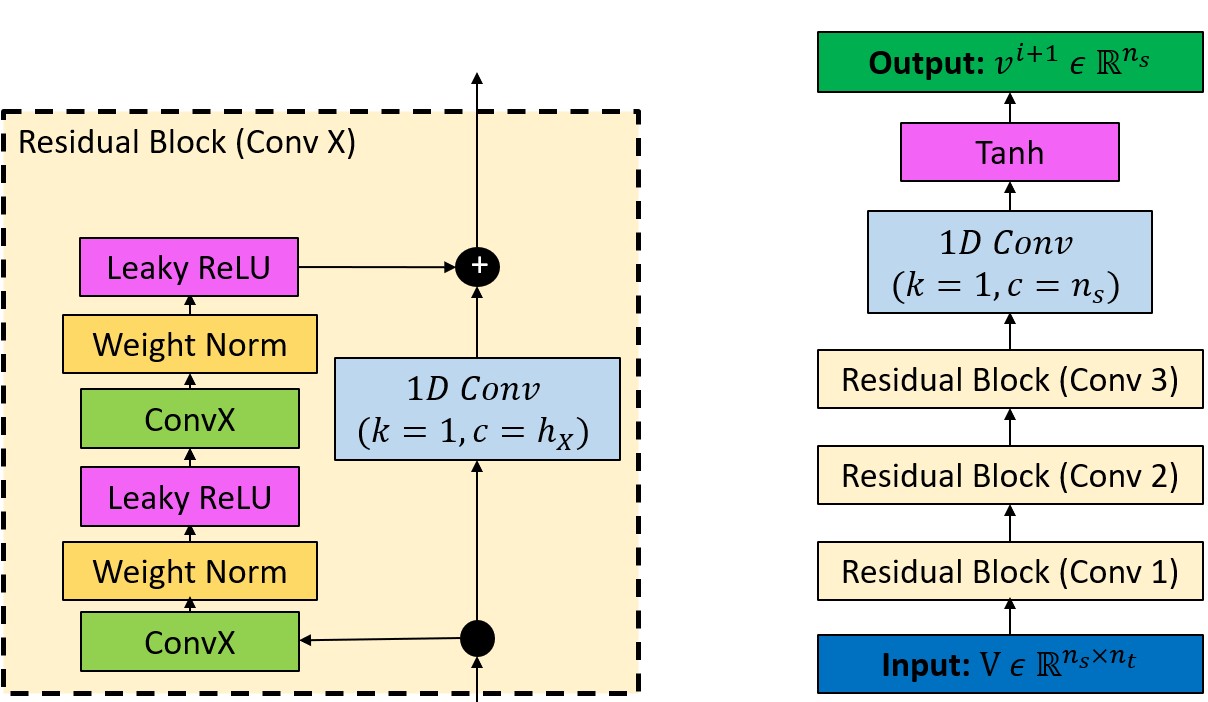}
    \caption{Residual Block (left) and TCN architecture (right)}
    \label{fig:8}
\end{figure}
The TCN is based on two principles \cite{BaiTCN2018}:  The network produces an output of the same length as the input, and  there can be no leakage from the future into the past. To verify that the first principle is respected, the TCN uses a 1D fully-convolutional network (FCN) where each hidden layer has the same length as the input layer, and zero padding of length $(k -1)$ is added to keep subsequent layers the same length as previous ones. To respect the second principle, the TCN uses causal convolutions (achieved by padding only on the starting side of input sequences), where the output at time $i$ is convolved only with elements from time $i$ and earlier in the previous layer (Figure \ref{fig:7}). A TCN also makes use of dilated convolutions that enable an exponentially large receptive field. For an input sequence, $V \ \epsilon$ $\mathbb{R}^{n_s \times n_t}$ and a kernel $K$ with learnable weights, $K \ \epsilon \  \mathbb{R}^k$ ($k$ is the kernel size), the element $O(s)$ with s $\epsilon \{0, 1, ..., n_t-k+1\}$ produced by the dilated 1D convolution is:

\begin{equation}
    O(s) = \sum_{j=0}^{k-1}V(s+j*d) \times K(j)
\end{equation}

\noindent where $d$ is the dilation factor and $k$ is the kernel size. When using dilated convolutions, $d$ is increased  exponentially with the depth of the network (eg., $d = 2^l$ at level $l$ of the network), ensuring that some filter hits each input within a large effective history.

In the TCN model employed here, a generic residual block is used in place of a convolutional layer. A residual block contains a branch leading to a series of transformations obtained by layers of TCNs, whose outputs are added to the input $V$ of the block to obtain $O_{rb}$:

\begin{equation}
    O_{rb} = Activation(V + F(V))
\end{equation}

\noindent Within a residual block (Figure \ref{fig:8}), the TCN has two layers of dilated causal convolution with weight normalization and non-linearity, with a leaky rectified linear unit (leaky ReLU). To account for different input-output widths during addition operations, a 1D convolution (kernel size = 1 and channels = $n_s$) is used to ensure the element-wise addition operator ($\oplus$) receives tensors of the same shape.

When convolving along the temporal axis, this ({\it standard}) TCN model uses information available from all the prior time-steps (due to the large receptive field) to evaluate the next time-step, as sketched in Figure \ref{fig:7}. The model takes in a sequence of $n_{t}$ vectors corresponding to a look-back window of size $n_{t}$ : $V = [v^{i-n_{t}+1}, …, v^{i-1}, v^{i}]$, with $V \,  \epsilon \, \mathbb{R}^{n_s \times n_t}$. The filters convolve along the temporal axis for all the $n_s$ vector nodes, since the nodes are passed in as the channels. However, the results produced from this model (Section \ref{sec: Results}) do not propagate beyond the training domain. Therefore, another model is proposed here, where the dilated convolutions of the TCN model convolve along the spatial axis and thus use the information available from the neighbouring nodes to determine the future time-step value of the node. This model takes in a sequence of $n_{t}$ vectors corresponding to a look-back window of size $n_{t}$ in a transposed manner, such that the $n_{t}$ solution vectors are on separate channels: $V^T \,  \epsilon \, \mathbb{R}^{n_t \times n_s}$, where $V = [v^{i-n_{t}+1}, …, v^{i-1}, v^{i}]$. This model produces significantly better results than the TCN on a temporal axis, but the causal padding and dilations employed are of no significance when the convolution filter operates along the spatial axis. Another architecture for modeling the system dynamics, with 1D convolutions and without any dilations or causal paddings, is therefore proposed in the following section.

\subsubsection{ A Convolution Neural Network (CNN) for time forecasting}
\label{subsubsec:CNN}

A convolutional layer convolves filters with trainable weights on the input vector $v^i$ \cite{Goodfellow-et-al-2016}. Such filters are commonly referred to as convolutional kernels. In a convolutional neural network, the inputs and outputs can have multiple channels. For a convolutional layer with $n_{in}$ input channels and $n_{out}$ output channels, the total number of convolutional kernels is $n_{k} = n_{i} \times n_{o}$. Each kernel slides along the spatial direction, and the products of kernel weights and vector nodes are computed at all sliding steps. For an input vector $v^i$ and a  kernel $K$, the corresponding output feature map $O(s)$ with s $\epsilon \{0, 1, ..., n_s-k+1\}$ (where $k$ is the kernel size) is given by:

\begin{equation}
    O(s) = \sum_{j=0}^{k-1}v^i(s+j) \times K(j)
\end{equation}

\begin{figure}[t!]
    \centering
    \includegraphics[width = \textwidth]{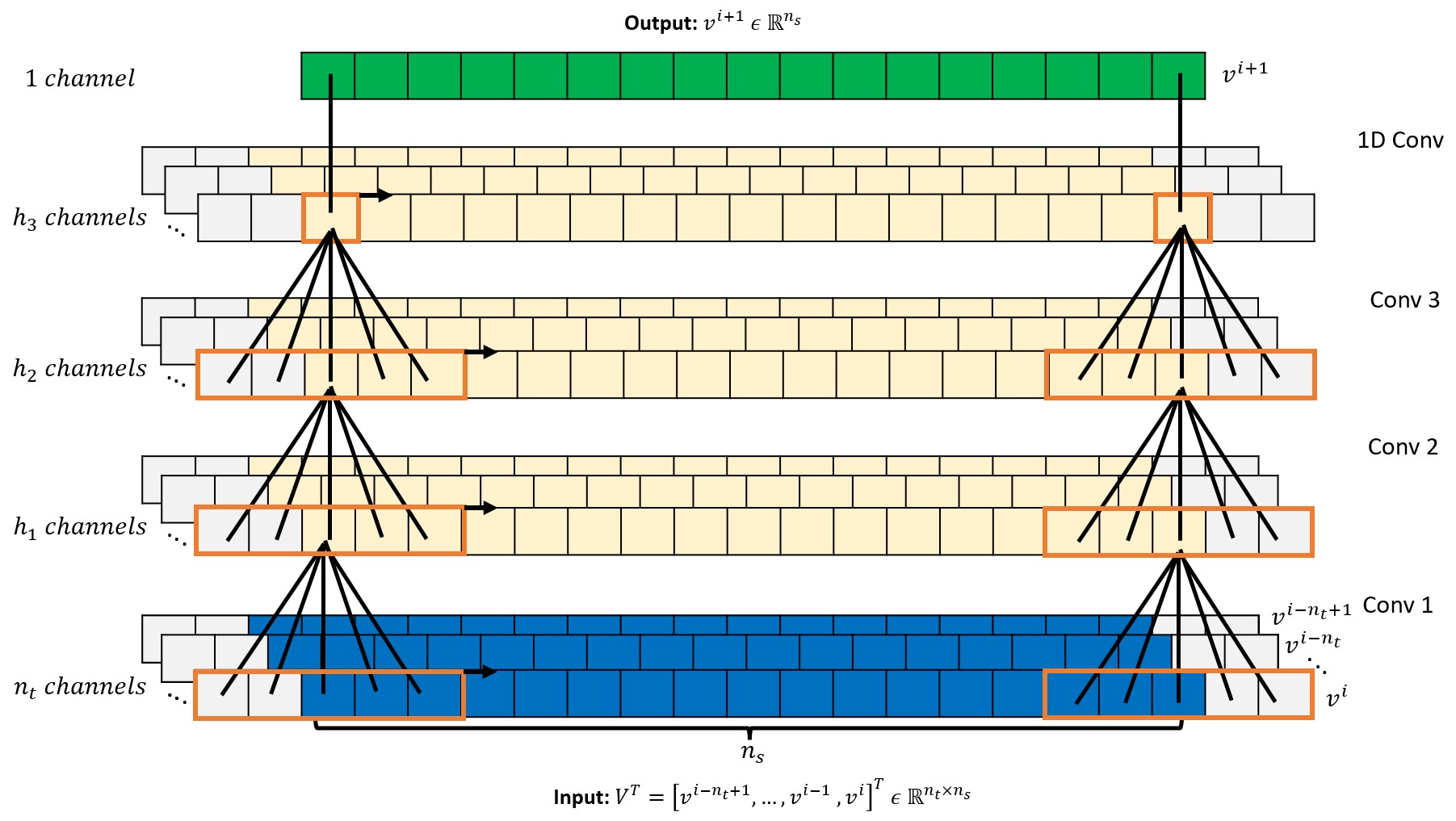}
    \caption{1D CNN filters convolving on spatial dimension of vectors}
    \label{fig:9}
\end{figure}

\begin{figure}[t!]
    \centering
    \includegraphics[width = \textwidth]{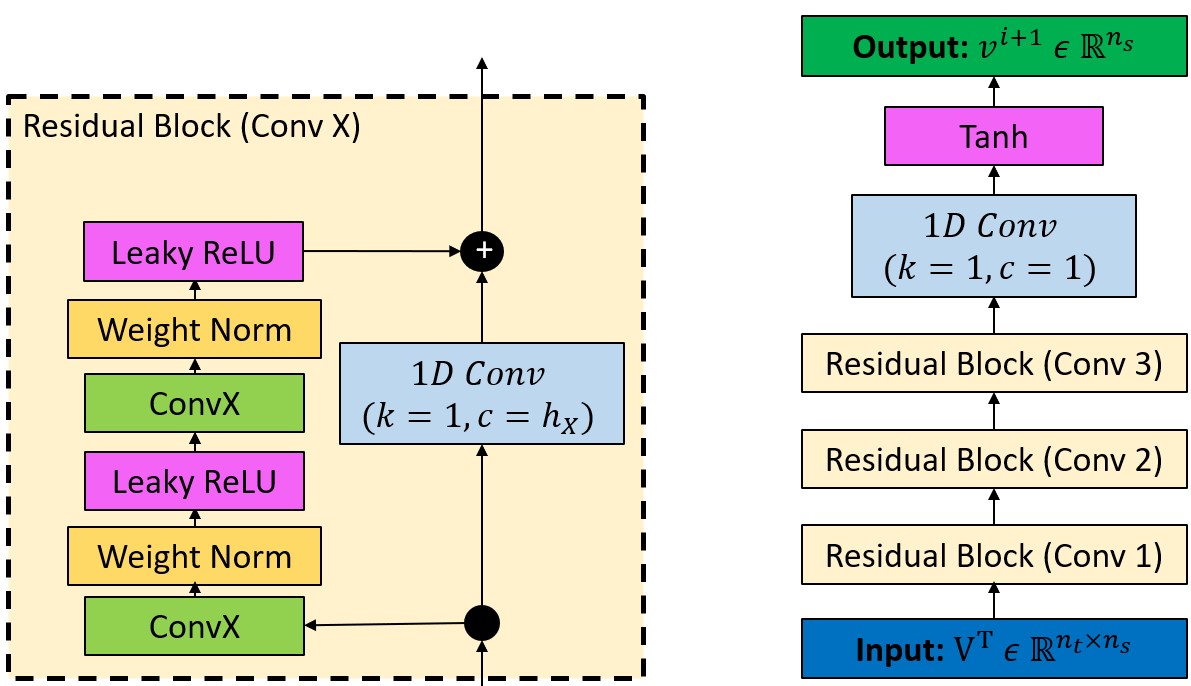}
    \caption{Residual Block (left) and CNN architecture (right)}
    \label{fig:10}
\end{figure}

\noindent Zero padding of size $(k-1)/2$ is added to both sides of the output feature map to maintain the spatial dimension as $n_s$. The forecasting model of CNN takes in a sequence of vectors with $n_{t}$ time-steps in a transposed manner as its input: $V^T \,  \epsilon \, \mathbb{R}^{n_t \times n_s}$, where $V = [v^{i-n_{t}+1}, …, v^{i-1}, v^{i}]$, so that the filter convolves on the spatial dimension of size $n_s$, and the $n_{t}$ vectors lie on separate channels, as shown in Figure \ref{fig:9}. The CNN architecture (Figure \ref{fig:10}) consists of $X$ residual blocks ($X$ is a hyperparameter), in which the input to each block, after transformation (to make the channels equal) from a 1D Convolution layer (kernel = 1 and channels = 1) is added to the output from the block. A residual block consists of two convolution layers, each followed by a weight normalization and a leaky ReLU activation layer.

\subsection{Uncertainty Quantification using Deep Ensembles}
\label{subsec:ensemble}

Deep neural networks, when applied in their traditional form, only predict the mean values of the output and do not provide any information regarding the uncertainty in the predicted output. Deep Ensembles address this issue by using an ensemble of variance-informed deep neural networks. Such neural networks possess a dual output \cite{jacquier_abdedou_delmas_soulaimani_2021,nix_weigend_1994, Lakshminarayanan2017SimpleAS}. In the context of the forecasting models employed with deep ensembles in this study, this means that the output size is twice the dimension $m$ of the predicted latent vectors ($z^{i+1}$) of solution steps, because the output contains both a mean value $\mu^z$ and a raw variance $\rho^z$. The raw variance is constrained to positiveness via a softplus activation, which produces the output variance ${\sigma^{z}}^2$.

\begin{equation}
    {\sigma^{z}}^2 = softplus(\rho^z) = \log(1+\exp{\rho^z})
\end{equation}

The predicted variance accounts for the noise or spread in the data utilized by the forecasting models, thereby becoming a measure of the epistemic uncertainty \cite{10.5555/3295222.3295309}.

\subsubsection{Ensemble model training}

The post-compression dataset (Section \ref{subsec:Dataset})  from the encoder network produces $N$ samples of the form: {$Z\epsilon \, \mathbb{R}^{m \times n_t}, \, z^{i+1} \epsilon \, \mathbb{R}^m$}, which train the forecasting model with the objective of minimizing the negative log-likelihood (NLL) loss function, as given below:

\begin{equation}
J_{NLL}(\mathbf{\theta}) = \frac{1}{N}\sum_{i=1}^N \frac{\log{{\sigma^z_{i}}^2}}{2}+\frac{(z_i-\mu^z_{i})^2}{2{\sigma^z_{i}}^2}
\end{equation}

An optimizer (Adam algorithm \cite{https://doi.org/10.48550/arxiv.1412.6980}) is used to update the model parameters (weights and biases) - $\mathbf{\theta (w, b)}$. To account for the epistemic uncertainty, $M$ sets of $\mathbf{\theta (w, b)}$  are randomly initialized, thereby creating $M$ independent forecasting networks and introducing variability in the training step \cite{Lakshminarayanan2017SimpleAS, NEURIPS2019_8558cb40}. The epistemic uncertainty is associated with the model variance and its data-fitting capabilities, which diminish with the increase in training data. This uncertainty is more dominant in our case, compared to aleatoric uncertainty, since the dataset is obtained by analytical solutions. Therefore, the variance behaves as an indicator of the likelihood that the model is making predictions that are out-of-distribution.

The predicted latent moments ($\mu^z_{\theta}$, ${\sigma^z_{\theta}}^2$) from each model create a probability mixture that can be approximated as a single Gaussian distribution, leading to a mean and variance  obtained for the ensemble as:

\begin{equation}
\mu^z_* = \frac{1}{M}\sum_{i=1}^M \mu_{\theta^z_i}
\end{equation}
\noindent and
\begin{equation}
{\sigma^z_*}^2 = \frac{1}{M} \sum_{i=1}^M  ({\sigma^z_{\theta_i}}^2+\mu^z_{\theta_i}) - \mu^z_*
\end{equation}

\subsubsection{Predictions of the expanded solution vectors}

During the online (testing) stage, the forecasting model predicts the latent vector $z^i$ along with the confidence interval at all  time-steps after the input sequence. To obtain the full physical solution, the decoder ($\chi_d$) is used to reconstruct or expand the obtained latent vector. Although the decoding operation can be directly applied to the predicted mean $\mu^z_*$, this does not hold true for the predicted variance ${\sigma^z_*}^2$, because it does not transform linearly to the expanded space by using the decoding operation. Since the latent vector $z^i$ is represented as a normal distribution $\hat{z}^i \sim \mathbb{N}(\mu^z_*, {\sigma^z_*}^2)$, a sufficiently large sample from $\hat{z}^i$ can be drawn and reconstructed individually to produce the full solution vector $\hat{v}^i=\chi_d(\hat{z}^i)$.

The unscented transform, as proposed by Julier et. al \cite{julier_uhlmann_1997}, is used to calculate the predicted mean ($\mu^v_*$) and variance (${\sigma^v_*}^2$) of the full-order solution in expanded space ($v^i \, \epsilon \, \mathbb{R}^{n_s}$) from the predicted mean and variance of $z^i$, with the help of nonlinear transformation ($\chi_d$). A set of $2m+1$ points (known as sigma points) with a sample mean and sample variance, $\mu^z_*$  and ${\sigma^z_*}^2$, respectively, are chosen, and the nonlinear function is applied to each point to yield a set of transformed points, of   $\mu^v_*$  and ${\sigma^v_*}^2$ sample mean and variance, respectively. The primary difference of this approach from the Monte-Carlo method is that the samples are not drawn at random but instead according to a specific, deterministic algorithm. The points are obtained using:

\begin{equation}
\begin{aligned}
    \tilde{z_0} = \mu^v_* , \, \, \,  W_0 = k/(m+k) \\
    \tilde{z_i} = \mu^v_*+(\sqrt{(m+k)P_{zz}})_i , \, \, \, W_i = 1/2(m+k) \\
    \tilde{z}_{i+n} = \mu^v_*-(\sqrt{(m+k)P_{zz}})_i , \, \, \, W_i = 1/2(m+k)
\end{aligned}
\label{eq14}
\end{equation}

\noindent where $k \, \epsilon \, \mathbb{R}$ is a constant, $P_{zz} \, \epsilon \, \mathbb{R}_{m \times m} $ is the covariance matrix formed by placing the elements of ${\sigma^z_*}^2$ as the diagonal, $(\sqrt{(m+k)P_{zz}})_i$ is $i^{th}$ row or column of the matrix square root of $(m+k)P_{zz}$ and $W_i$ is the weight associated with the $i^{th}$ point. The transformation is then achieved via the following procedure: 

1. Transform each sigma point ($i \, \epsilon \{0, 1, ..., 2n\}$) by the non-linear decoder function $\chi_d$ to obtain:
\begin{equation}
  y_i = \chi_d(\tilde{z_i})
  \label{eq15}
\end{equation}

2. Predict the mean for the full-order solution using the weighted average of transformed points:
\begin{equation}
 \mu^v_* = \sum_{i=0}^{2n} W_i y_i   
\end{equation}
3. Predict the variance for the full order solution from the diagonal elements of the covariance matrix $P_{vv}$, which are the weighted outer products of the transformed points:
\begin{equation}
     P_{vv} = \sum_{i=0}^{2n} W_i(y_i-\mu^v_*)(y_i-\mu^v_*)^T \\
\end{equation}
\begin{equation}
    {\sigma^v_*}^2 = diag(P_{vv})
\end{equation}

\subsubsection{Metrics}

To evaluate the performance of the previous architectures, the following metrics are used:

\noindent {\it{Mean Squared Error ($L_2$ Norm)}}: 
The average of the square of the difference between the actual $v_i$ and predicted values $\hat v_i$ over $N$ samples:

\begin{equation}
    MSE = \frac{\sum_{i=1}^N(v_{i}-\hat v_{i})^2}{N}
\end{equation}

\noindent {\it{Mean Absolute Error ($L_1$ Norm)}}:
The average of the difference between the two vectors $v_i$ and $\hat v_i$ over $N$ samples:

\begin{equation}
    MAE = \frac{\sum_{i=1}^N \| v_{i}-\hat v_{i}\|} {N} 
\end{equation}

\noindent {\it{Relative $L_2$ Norm Error}}:
The relative $L_2$ norm error (referred as error) is calculated  as:

\begin{equation}
    Relative Error = \frac{\sqrt{\sum_{i=1}^N (v_{i}-\hat v_{i})^2}}{\sqrt{\sum_{i=1}^N v_{i}^2}}
\end{equation}

\section{Results and Discussion}
\label{sec:Results}

The capability of the autoencoders (MLP-AE and CAE) to efficiently transform high-dimensional vectors to a low dimensional space, and that of the forecasting models (LSTM, TCN, and CNN) to accurately model the system dynamics were tested using advection-dominated flow problems (1D Burgers' and Stoker's problems).

\subsection{1D Burgers' problem}

The test case involves the one-dimensional Burgers’ equation, which is a non-linear advection-diffusion PDE. The equation along with the initial and Dirichlet boundary conditions are given by

\begin{equation}
\frac{\partial u}{\partial t} + u\frac{\partial u}{\partial x} = \nu\frac{\partial^2 u}{\partial t^2}
\end{equation}
\begin{equation}
 u(x, 0) = u_0, x \epsilon [0, L], u(0, t) = u(L, t) = 0 \end{equation}
\begin{equation}
u(x, 0) \equiv u_0 = \frac{x}{1+\sqrt{\frac{1}{t_0}}\exp({Re\frac{x^2}{4}})}
\end{equation}
		 
\noindent where the length $L = 1m$ and the maximum time $T_{max} = 2s$. The solutions obtained from the above equations produce sharp gradients even with smooth initial conditions if the viscosity $\nu$ is sufficiently small, due to the advection-dominated behavior. The analytical solution of the problem is given by:

\begin{equation}
    u(x, t) = \frac{\frac{x}{t+1}}{1+\sqrt{\frac{t+1}{t_0}}\exp({Re\frac{x^2}{4t+4}})}
\end{equation}
 
\noindent where $t_0 = \exp({\frac{Re}{8}}) $ and $Re = 1/\nu$. The high-fidelity solution vectors are generated by directly evaluating the analytical solution over a uniformly discretized spatial domain containing 200 grid points ($n_s = 200$) at 250 uniform time-steps ($T = 250$) for two different values of $Re$: 300 and 600. The solution vectors obtained are then used to train the autoencoder and forecasting models (Section \ref{subsec:Dataset}). For  the autoencoder training, 200 solution vectors are chosen at random time-steps, and the remaining 50 are used for validation. For the forecasting model, the training set is comprised of the first 150 compressed samples, each sample containing $n_t$ consecutive solution vectors (i.e. look back window = $n_t$), where $n_t$ is a hyperparameter. The validation set is composed of the remaining samples. For testing, $n_t$ latent vectors from the start of the dataset are fed to the forecasting model to predict the subsequent time-steps via auto-regression (Table \ref{tab:table1}).

\begin{table}[h!]
    \centering
    \begin{tabular}{c c c c}
         \hline
         Dataset & Samples & Input & Output\\ [0.5ex] 
         \hline\hline
         
         \multirow{4}{*}{Training} & 1 & $[z^{1},…, z^{n_t-1}, z^{n_t}]$ & $z^{n_t+1}$ \\
         & 2 & $[z^{2},…, z^{n_t}, z^{n_t+1}]$ & $z^{n_t+2}$\\
         & ... & ... & ...\\
         & 150 & $[z^{150},…, z^{n_t+148}, z^{n_t+149}]$ & $z^{n_t+150}$ (training end)\\
         \hline
        
        \multirow{3}{*}{Validation} & 151 & $[z^{151},…, z^{n_t+149}, z^{n_t+150}]$ & $z^{n_t+151}$ \\
         & ... & ... & ...\\
         & 250-$n_t$ & $[z^{250-n_t},…, z^{248}, z^{249}]$ & $z^{250}$\\
         \hline
         Testing & 1 & $[z^{1},…, z^{n_t-1}, z^{n_t}]$ & $[z^{n_t+1},…, z^{249}, z^{250}]$ \\
         \hline
    \end{tabular}
    \caption{Burgers' problem, Training, Validation and Testing dataset}
    \label{tab:table1}
\end{table}

\subsubsection{Autoencoders for spatial compression}
Two types of autoencoder architectures (Section \ref{subsec:NIROM}) - MLP (referred to as AE) and Convolutional (referred to as CAE) are proposed for the compression of  solution vectors - $v^i \, \epsilon \mathbb{R}^{n_s}$ to latent vectors $z^i \, \epsilon \mathbb{R}^{m}$ by the encoder $\chi_e$. Sequences formed from these latent vectors ($[z^{i-n_{t}+1}, …, z^{i-1}, z^{i}] \,  \epsilon \, \mathbb{R}^{m \times n_t}$) are  utilized to train the forecasting models - LSTM, TCN and CNN. The trained models are then used to forecast the latent vectors at subsequent time-steps to the sequence ($[z^{n_{t}+1}, z^{n_{t}+2}, …, z^{T}] \,  \epsilon \, \mathbb{R}^{m \times (T-n_t)}$) given as input to the forecasting model. The latent vectors are then reconstructed into solution vectors ($[v^{n_{t}+1}, v^{n_{t}+2}, …, v^{T}] \,  \epsilon \, \mathbb{R}^{n_s \times (T-n_t)}$) using the decoder $\chi_d$. The heat map plots obtained by stacking these reconstructed solution vectors along the x-axis (spatial nodes-$n_s$ along y, time-steps-$n_t$ along x), are illustrated for both autoencoder models in Appendix A (Table \ref{tab:A.1}). 

Both architectures, AE and CAE,  are capable of efficiently compressing the solution vectors to latent vectors with few modes and fine reconstruction/decompression. However, only CAE compression followed by CNN autoregression produces accurate results on extrapolation. This is because the proposed CAE architecture is devoid of any dense layer (single layer of neurons), and therefore even during compression, the local spatial information in the vector remains preserved. This consistency facilitates the modeling of latent dynamics by the CNN model, as it convolves on the spatial axis of the input and utilizes information from the neighboring cells at the provided time-steps to predict nodal values at subsequent time-steps.  The hyperparameters for the AE and CAE architectures are listed in Table \ref{tab:table2}, where encoder layers denote the number of neurons in the two dense encoder layers of AE, and the number of channels in the convolution layers of CAE. The decoders of both autoencoders are mirrored structures of their encoders.

\begin{table}[h]
    \centering
    \begin{tabular}{c c c}
         \hline
         Hyperparameters & MLP AE & Convolutional CAE\\ [0.5ex] 
         \hline\hline
         Encoder layers & [100, 50] & [8, 32] \\ 
         \hline
         latent dimension ($m$) & 10, 25, 50 & 12, 25, 50 \\
         \hline
         Activation & relu, swish &relu, swish\\
         \hline
         Loss Function & MSE & MSE \\
         \hline
         Learning rate & $10^{-3},3 \times 10^{-4}$& $10^{-3},3 \times 10^{-4}$\\
         \hline
    \end{tabular}
    \caption{Hyperparameters for the AE and CAE networks}
    \label{tab:table2}
\end{table}

\subsubsection{LSTM model}

When LSTM (Section \ref{subsubsec:LSTM})is used as the future step predictor, it takes in a sequence of $n_t$ (lookback window) latent vectors (spatial dimension = $m$) obtained by compression from the encoder network ($Z \,  \epsilon \, \mathbb{R}^{m \times n_t}$) to produce the latent vector for the next time-step ($z^{i+1}\, \epsilon \, \mathbb{R}^{m} $). The LSTM model consists of multiple LSTM layers stacked together, each having a hidden dimension equal to the latent dimension of the solution vectors. Various sets of  hyperparameters considered for the LSTM network, for both Re 300 and 600 are summarized in Table \ref{tab:table3}:
\begin{figure}[t!]
    \centering
    \begin{subfigure}{\textwidth}
    \begin{subfigure}{0.33 \textwidth}
    \centering
    \includegraphics[width = \textwidth]{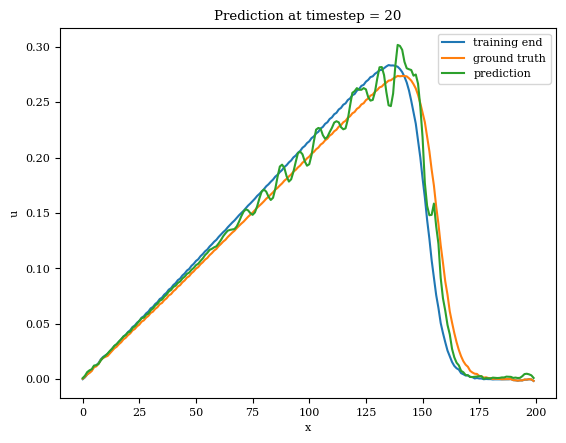}
    \end{subfigure}%
    \begin{subfigure}{0.33 \textwidth}
    \centering
    \includegraphics[width = \textwidth]{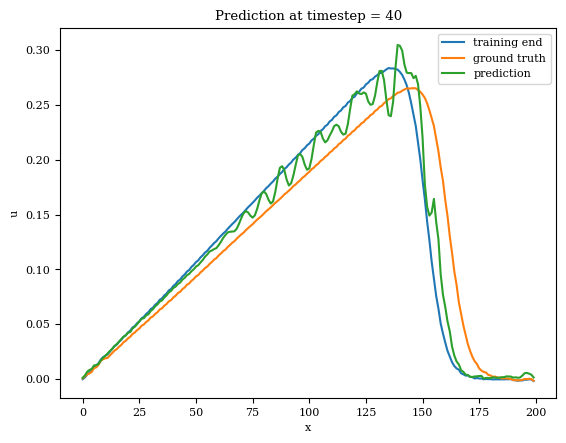}
    \end{subfigure}%
    \begin{subfigure}{0.33 \textwidth}
    \centering
    \includegraphics[width = \textwidth]{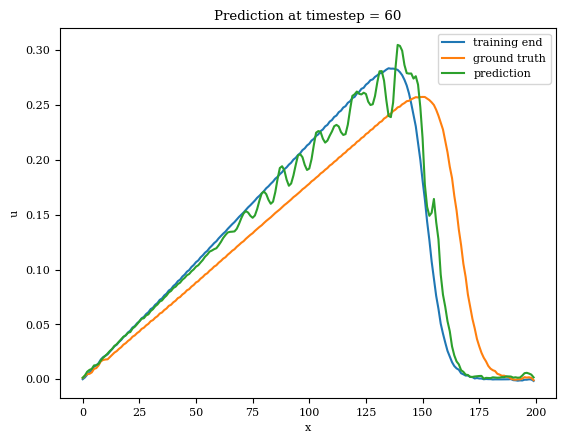}
    \end{subfigure}%
    \caption{Re = 300}
    \end{subfigure}%
    \\
    \begin{subfigure}{\textwidth}
    \begin{subfigure}{0.33 \textwidth}
    \centering
    \includegraphics[width = \textwidth]{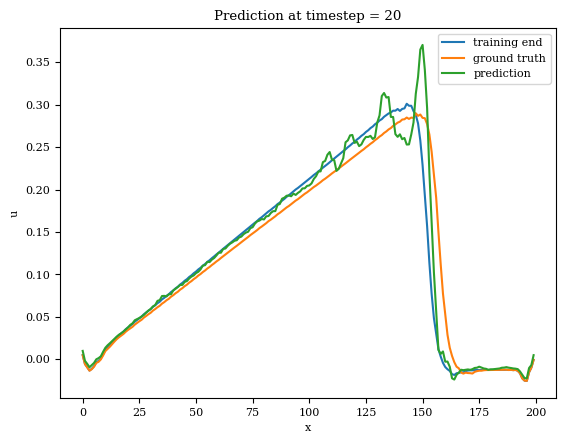}
    \end{subfigure}%
    \begin{subfigure}{0.33 \textwidth}
    \centering
    \includegraphics[width = \textwidth]{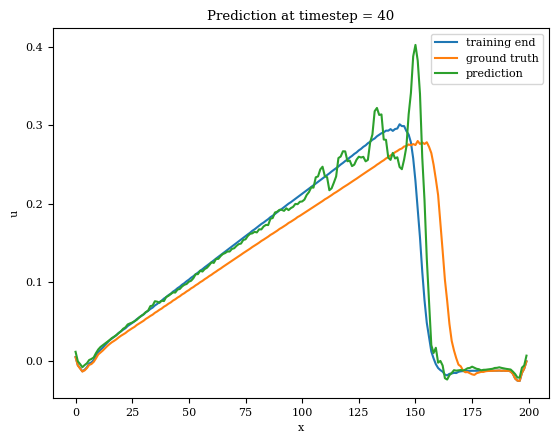}
    \end{subfigure}%
    \begin{subfigure}{0.33 \textwidth}
    \centering
    \includegraphics[width = \textwidth]{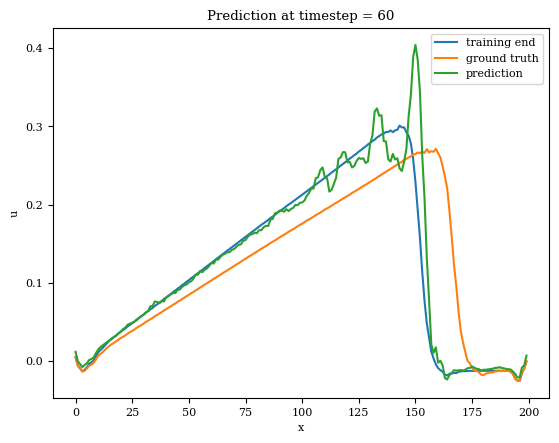}
    \end{subfigure}%
    \caption{Re = 600}
    \end{subfigure}%
    \caption{Burgers' problem, Extrapolative auto-regressive predictions  by the LSTM modelfor time-steps = 180, 200 and 220. The training end time-step = 160; for Re=300 (a) and Re= 600 (b)}
    
    \label{fig:11}
\end{figure}

\begin{figure}[t!]
    \centering
    \centering
    \begin{subfigure}{.5 \textwidth}
    \centering
    \includegraphics[width = \textwidth]{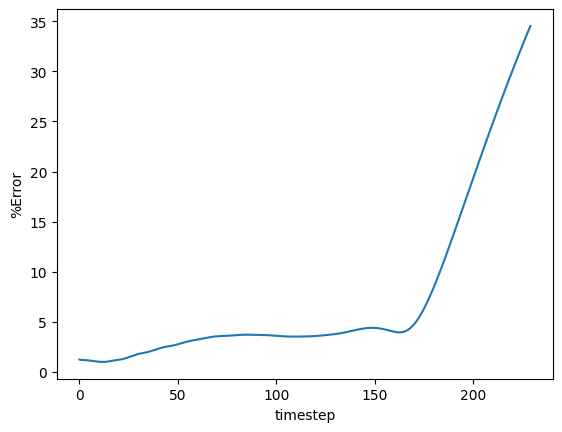}
    \caption{Re = 300}
    \end{subfigure}%
    \begin{subfigure}{.5 \textwidth}
    \centering
    \includegraphics[width = \textwidth]{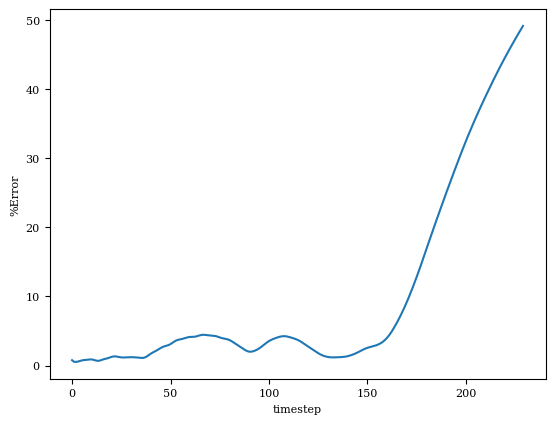}
    \caption{Re = 600}
    \end{subfigure}
    \caption{Burgers' problem, $L_2$ relative error of the autoregressive predictions with increasing time for Re = 300 (a) and Re = 600 (b) for the LSTM model}
    \label{fig:12}
\end{figure}
\begin{table}[h]
    \centering
    \begin{tabular}{c c c}
         \hline
         Hyperparameters & Values\\ [0.5ex] 
         \hline\hline
         Sequence length ($n_t$) & 5, 10, 20\\
         \hline
         LSTM layers & 1, 2, 3\\ 
         \hline
         hidden/latent dimension ($m$) & 12,25,50 \\
         \hline
         Activation & tanh\\
         \hline
         Loss Function & MSE\\
         \hline
         Learning rate & $5 \times 10^{-4}$\\
         \hline
    \end{tabular}
    \caption{Hyperparameters for the LSTM network}
    \label{tab:table3}
\end{table}

The models are trained in batches of size 15, and the loss values for both training and validation converge in 3000 epochs. The model with the least validation loss has a lookback window of size 10 and a single LSTM layer with hidden dimension 50 for both Re 300 and 600. The extrapolation (Figure \ref{fig:11}) and error plots (Figure \ref{fig:12}) obtained from these models show that the LSTM model accurately predicts the solution vectors for time-steps within the training domain ($i <= 150$), but the solution does not change for time-steps outside the training domain, and so  the relative error increases drastically, reaching 35\% for Re= 300 and 50\% for Re=600.

\subsubsection{TCN model}

Similar to the LSTM, the TCN model (Section \ref{subsubsec:TCN}) takes in a sequence of $n_t$ latent vectors obtained by compression from the encoder network ($Z \, \epsilon \, \mathbb{R}^{m \times n_t}$) to produce the latent vector for the next time-step ($z^{i+1}\, \epsilon \, \mathbb{R}^{m} $), with the dilated convolutions operating on the temporal axis, and the latent dimension passed as a channel. The TCN model consists of either 2 or 3 TCN blocks, each with the same kernel size and number of channels, but with dilations increasing by a factor of 2 in subsequent blocks. The hyperparameters for the TCN network for both Re 300 and 600 are summarized in Table \ref{tab:table4}.

\begin{table}[h]
    \centering
    \begin{tabular}{c c c}
         \hline
         Hyperparameters & Values\\ [0.5ex] 
         \hline\hline
         Sequence length ($n_t$) & 5, 10, 20\\
         \hline
         TCN block channels & [32, 32], [64, 64], [32, 32, 32], [64,64,64]\\ 
         \hline
         latent dimension ($m$) & 12, 25, 50 \\
         \hline
         Kernel Size(k) & 3, 5, 7, 9 \\
         \hline
         Activation & tanh\\
         \hline
         Loss Function & MSE\\
         \hline
         Learning rate & $1 \times 10^{-4}$\\
         \hline
    \end{tabular}
    \caption{Hyperparameters for the TCN network}
    \label{tab:table4}
\end{table}
\begin{figure}[t!]
    \centering
    \begin{subfigure}{\textwidth}
    \begin{subfigure}{0.33 \textwidth}
    \centering
    \includegraphics[width = \textwidth]{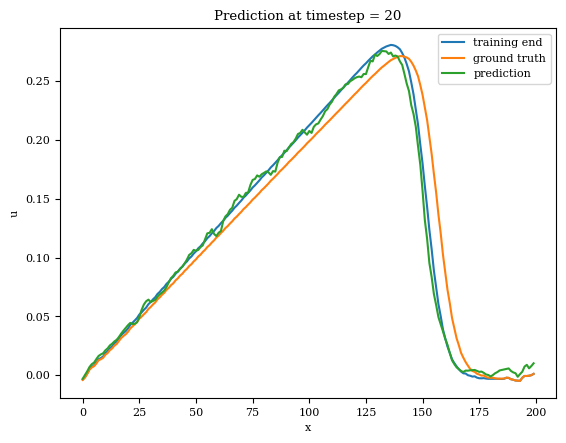}
    \end{subfigure}%
    \begin{subfigure}{0.33 \textwidth}
    \centering
    \includegraphics[width = \textwidth]{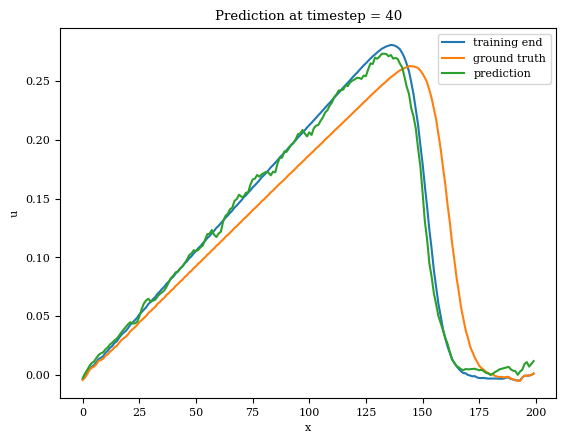}
    \end{subfigure}%
    \begin{subfigure}{0.33 \textwidth}
    \centering
    \includegraphics[width = \textwidth]{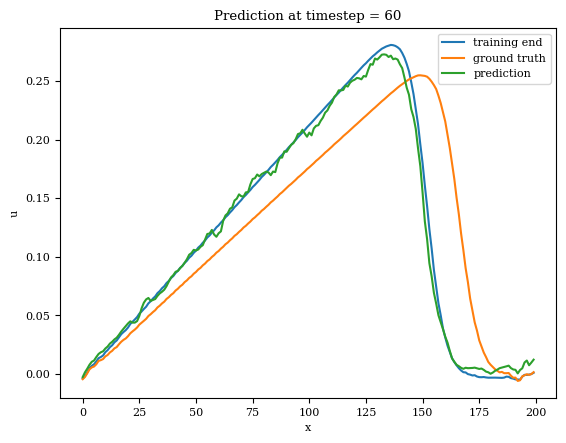}
    \end{subfigure}%
    \caption{Re = 300}
    \end{subfigure}%
    \\
    \begin{subfigure}{\textwidth}
    \begin{subfigure}{0.33 \textwidth}
    \centering
    \includegraphics[width = \textwidth]{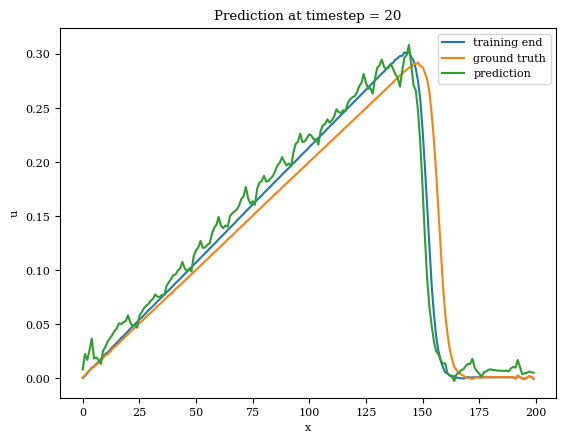}
    \end{subfigure}%
    \begin{subfigure}{0.33 \textwidth}
    \centering
    \includegraphics[width = \textwidth]{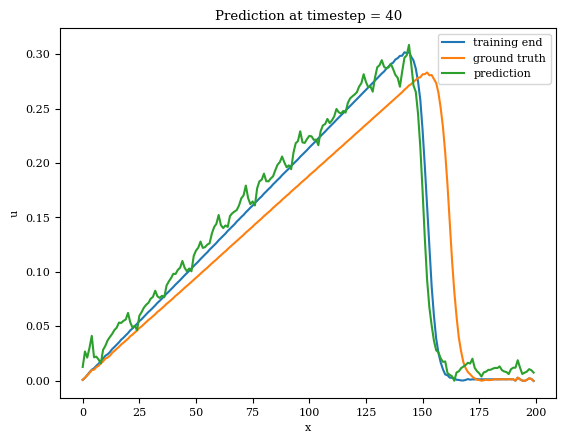}
    \end{subfigure}%
    \begin{subfigure}{0.33 \textwidth}
    \centering
    \includegraphics[width = \textwidth]{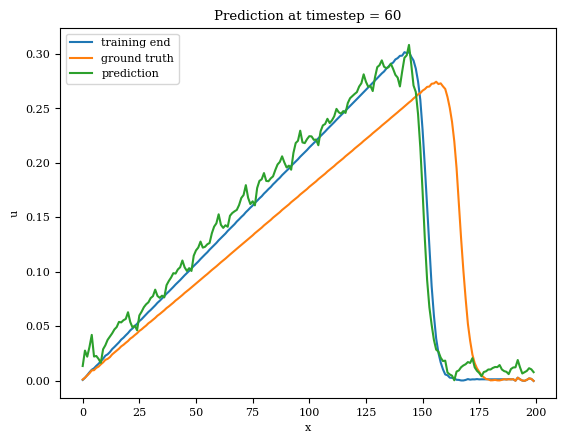}
    \end{subfigure}%
    \caption{Re = 600}
    \end{subfigure}%
    \caption{Burgers' problem, Extrapolative auto-regressive predictions using the TCN model (over time) for Re= 300 (a) and Re= 600 (b) and for time-steps = 180, 200 and 220; the training end time-step = 160}
    
    \label{fig:13}
\end{figure}

\begin{figure}[t!]
    \centering
    \centering
    \begin{subfigure}{.5 \textwidth}
    \centering
    \includegraphics[width = \textwidth]{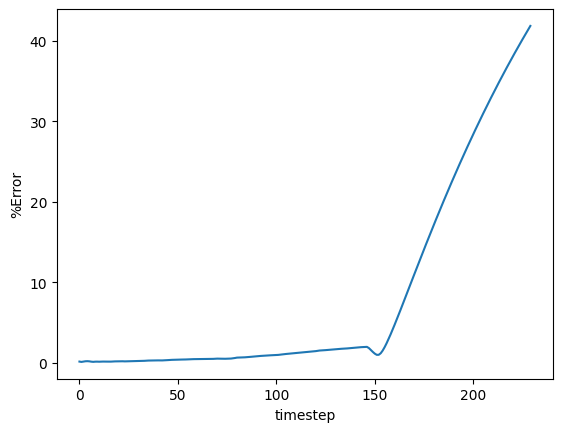}
    \caption{Re = 300}
    \end{subfigure}%
    \begin{subfigure}{.5 \textwidth}
    \centering
    \includegraphics[width = \textwidth]{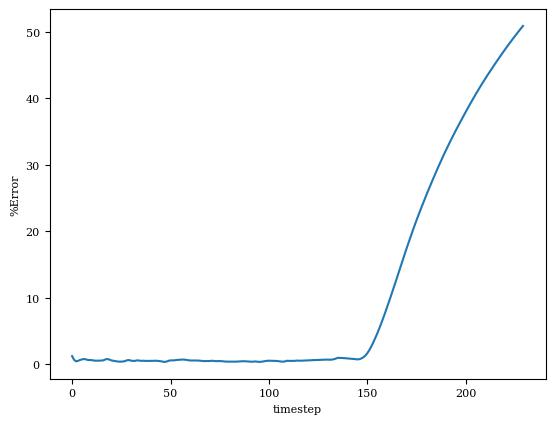}
    \caption{Re = 600}
    \end{subfigure}
    \caption{Burgers' problem, $L_2$  relative error of the auto-regressive predictions with increasing time for Re = 300 (a) and Re = 600 (b) for the TCN model (over time)}
    \label{fig:14}
\end{figure}
When models are trained in batches of size 15, the training and validation losses reach their minimum values in 4000 epochs. The model with the least validation loss takes in sequence with lookback window 10 and latent dimension 50. For Re 300, the best model has 3 temporal blocks, each having 64 channels, whereas for Re 600, it has 2 TCN blocks with kernel size 3 and 64 channels each. The extrapolation (Figure \ref{fig:13}) and error plots (Figure \ref{fig:14}) obtained from these models indicate that the TCN model accurately predicts the solution vectors for time-steps  within the training domain ($i <= 150$), but stops being accurate after the end of training, so that the error increases to 40\% for Re= 300 and 50\% for Re= 600. However, if the same model architecture operates on the input sequence, such that the dilated 1D convolutions propagate along the spatial axis, with each solution vector on a separate channel, then accurate forecasts are produced, even outside the training domain. This encourages the development of a simpler predictive/forecasting model, devoid of dilations and causal padding since the exponentially increasing receptive field serves no purpose when operating along the spatial axis.

\subsubsection{CNN model}
\begin{table}[b]
    \centering
    \begin{tabular}{c c c}
         \hline
         Hyperparameters & Values\\ [0.5ex] 
         \hline\hline
         Sequence length ($n_t$) & 5, 10, 20\\
         \hline
         CNN block channels & [50, 50], [100, 100], [200, 200]\\
         \hline
         latent dimension ($m$) & 12,25,50 \\
         \hline
         Kernel Size(k) & 3, 5, 7, 9 \\
         \hline
         Activation & tanh\\
         \hline
         Loss Function & MSE\\
         \hline
         Learning rate & $1 \times 10^{-4}$\\
         \hline
    \end{tabular}
    \caption{Hyperparameters for the CNN network}
    \label{tab:table5}
\end{table}

The proposed CNN model (Section \ref{subsubsec:CNN}) takes in a sequence of $n_t$ latent vectors  obtained by compression from the encoder network ($Z^T \, \epsilon \, \mathbb{R}^{n_t \times m}$) to produce the latent vector for the next time-step ($z^{i+1}\, \epsilon \, \mathbb{R}^{m} $), with 1D convolutions operating on the spatial axis (latent dimension) and $n_t$ latent vectors on separate channels. The CNN model consists of two residual blocks, each with the same kernel size and number of channels. The hyperparameters for the CNN network for both Re=300 and 600 are summarized in Table \ref{tab:table5}.

\begin{figure}[t!]
    \centering
    \begin{subfigure}{\textwidth}
    \begin{subfigure}{0.33 \textwidth}
    \centering
    \includegraphics[width = \textwidth]{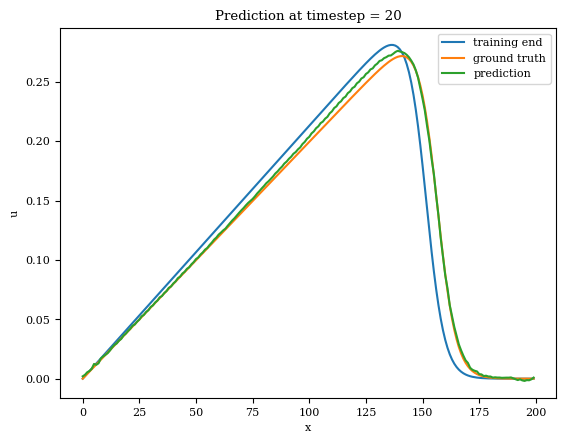}
    \end{subfigure}%
    \begin{subfigure}{0.33 \textwidth}
    \centering
    \includegraphics[width = \textwidth]{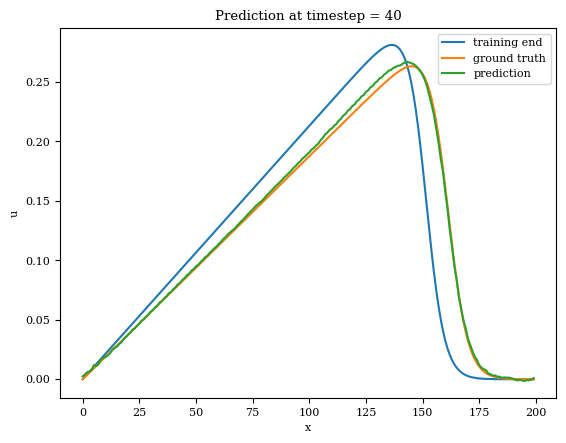}
    \end{subfigure}%
    \begin{subfigure}{0.33 \textwidth}
    \centering
    \includegraphics[width = \textwidth]{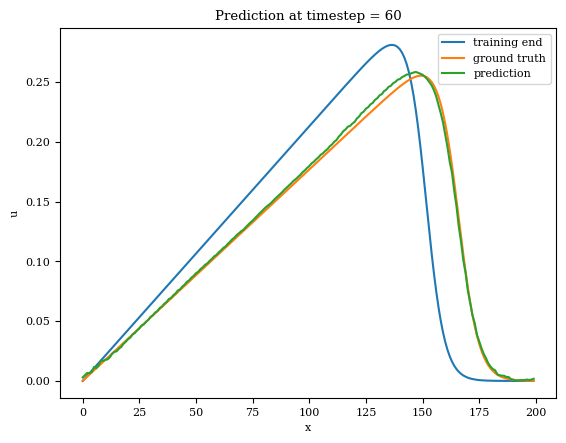}
    \end{subfigure}%
    \caption{Re = 300}
    \end{subfigure}%
    \\
    \begin{subfigure}{\textwidth}
    \begin{subfigure}{0.33 \textwidth}
    \centering
    \includegraphics[width = \textwidth]{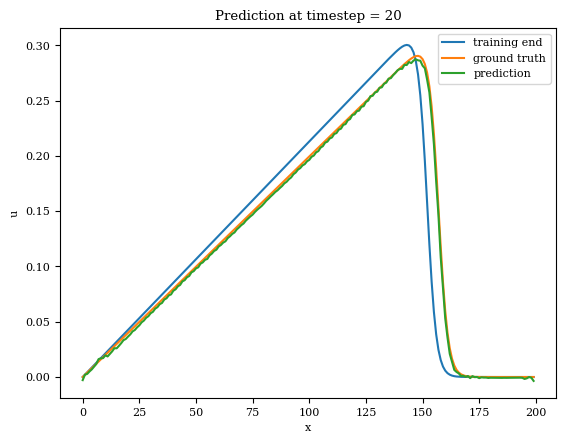}
    \end{subfigure}%
    \begin{subfigure}{0.33 \textwidth}
    \centering
    \includegraphics[width = \textwidth]{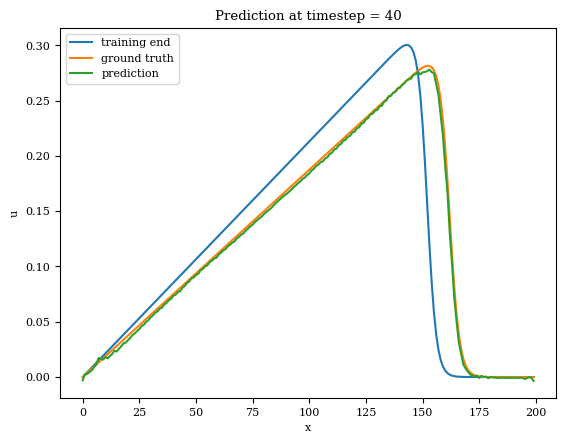}
    \end{subfigure}%
    \begin{subfigure}{0.33 \textwidth}
    \centering
    \includegraphics[width = \textwidth]{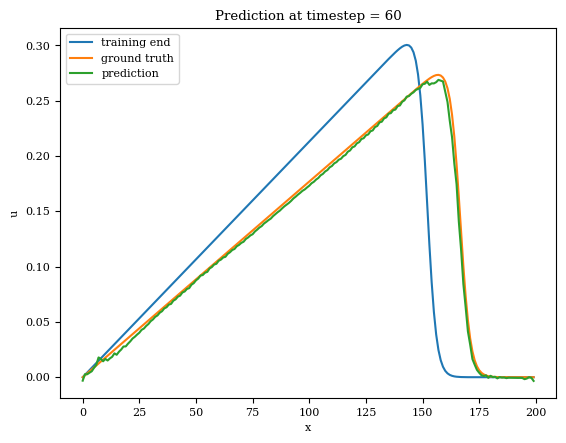}
    \end{subfigure}%
    \caption{Re = 600}
    \end{subfigure}%
    \caption{Burgers' problem: Extrapolative auto-regressive predictions using the proposed CNN model, for Re= 300 (a) and Re= 600 (b) and for time-steps = 180, 200 and 220; the training end time-step = 160}
    
    \label{fig:15}
\end{figure}

\begin{figure}[t!]
    \centering
    \centering
    \begin{subfigure}{.5 \textwidth}
    \centering
    \includegraphics[width = \textwidth]{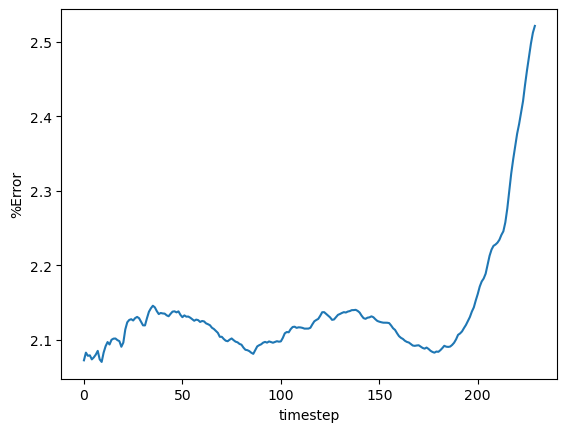}
    \caption{Re = 300}
    \end{subfigure}%
    \begin{subfigure}{.5 \textwidth}
    \centering
    \includegraphics[width = \textwidth]{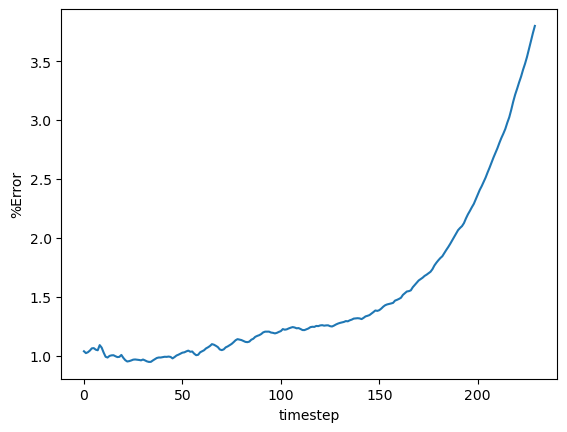}
    \caption{Re = 600}
    \end{subfigure}
    \caption{Burgers' problem: $L_2$  relative error of the auto-regressive predictions using the CNN model with increasing time for Re = 300 (a) and Re = 600 (b)}
    \label{fig:16}
\end{figure}

Training and validation loss converges in 3000 epochs for batch size 15. The model with the least validation loss has a lookback window of size 10, and each of its blocks has a kernel size of 3 and 50 channels for both Re= 300 and 600. It is clear from the extrapolation (Figure  \ref{fig:15}) and error plots (Figure \ref{fig:16}) that the CNN model accurately models the latent dynamics, and predicts solution vectors accurately for time-steps beyond the training domain. The error values increase with time, due to the accumulation of errors, since each subsequent time-step is predicted auto-regressively; i.e., using previously-predicted time-steps that contain slight errors. Still, the error reaches a mere 2.5\% for Re= 300 and 3.5\% for Re= 600, which is significantly less than that produced by other models.

\subsection{1D Stoker's Equation}

Stoker’s solution describes the propagation and rarefaction wave resulting from a one-dimensional dam break over a wet, flat, frictionless bottom. Stoker’s solution is considered among the most challenging benchmark test case due to its strong hyperbolic behavior and the discontinuity accompanying the propagation of the front wave resulting from the initial break. The dynamic is initiated by unequal water levels of both the upstream and downstream sides located in the middle of the studied domain of 100 m. 
 
The upstream water level is considered as an input random variable whose values are uniformly sampled within its plausible variability range $h_{up} \, \epsilon \, \mu [8, 11]$, whereas the downstream water depth is kept constant at a deterministic value $h_{ds} \, = \, 1 m$. The analytical solution for the water level is given as:

 \begin{equation}
     h(x, t) = \begin{cases}
    
     h_{up} & \text{if $x \leq x_A(t)$}\\
     \frac{4}{9g}(\sqrt{gh_{up}-\frac{x}{2t}})^2  & \text{if $x_A(t) \leq x \leq x_B(t)$}\\
     \frac{c_m^2}{g} & \text{if $x_B(t) \leq x \leq x_C(t)$}\\
     h_{ds}  & \text{if $x_C(t) \leq x$ }\\
    
     \end{cases} 
 \end{equation}

\begin{table}[b]
    \centering
    \begin{tabular}{c c c c}
         \hline\hline
         Dataset & Samples & Input & Output\\ [0.5ex] 
         \hline
         
         \multirow{4}{*}{Training} & 1 & $[z^{1},…, z^{n_t-1}, z^{n_t}]$ & $z^{n_t+1}$ \\
         & 2 & $[z^{2},…, z^{n_t}, z^{n_t+1}]$ & $z^{n_t+2}$\\
         & ... & ... & ...\\
         & 250 & $[z^{150},…, z^{n_t+248}, z^{n_t+249}]$ & $z^{n_t+250}$ (training end)\\
         \hline
        
        \multirow{3}{*}{Validation} & 251 & $[z^{251},…, z^{n_t+249}, z^{n_t+250}]$ & $z^{n_t+251}$ \\
         & ... & ... & ... \\
         & 450-$n_t$ & $[z^{450-n_t},…, z^{448}, z^{449}]$ & $z^{450}$\\
         \hline
         Testing & 1 & $[z^{1},…, z^{n_t-1}, z^{n_t}]$ & $[z^{n_t+1},…, z^{449}, z^{450}]$ \\
         \hline
    \end{tabular}
    \caption{Stoker's problem: Training, Validation and Testing dataset}
    \label{tab:table6}
\end{table}
\noindent where $x$ = the axial position, $x_A(t) = x_0-t\sqrt{gh_{up}}$, $x_B(t) = x_0 + t(\sqrt{gh_up}-3c_m)$ and $x_C = x_0+t\frac{2c_m^2(\sqrt{gh_up}-c_m)}{c_m^2-gh_{ds}}$, in which $c_m = \sqrt{gh_m} $  \cite{abdedou_stochastic}. For each selected value in the generated sample set of the upstream water level, the analytical solution given above is evaluated over 1000 nodes ($n_s = 1000$) that contain the computational domain $x \, \epsilon \, [0, 100]$ m for all 450 time-steps ($T = 450$) of the temporal domain $t \, \epsilon \, [0, 3.6] s$. Four-hundred solution vectors, at random time-steps, train the autoencoder network, and the remaining 50 vectors are used for the validation. The forecasting models are trained by the first 250 compressed samples and validated using the remaining 200. During testing, the first $n_t$ latent vectors are utilized to predict vectors at subsequent time-steps via auto-regression (Table \ref{tab:table6}).

\subsubsection{Autoencoder for spatial compression}
A similar methodology to the Burgers' test case \ref{sec:Results} is adopted for the training of autoencoder models, AE and CAE, and forecasting models, LSTM, TCN, and CNN, for the Stoker's problem. The heat map plots obtained by stacking the predicted solution vectors along the x-axis (spatial nodes-$n_s$ along y, time-steps-$n_t$ along x),  are illustrated for both autoencoder models in appendix A (Table \ref{tab:A.2}).

Both AE and CAE effectively transform the solution vectors to a reduced latent space, since they produce fine reconstruction for vectors within as well as outside of the training domain of the autoencoder model. But again, only the CAE-CNN model learns the latent dynamics accurately enough to predict solution vectors outside the training domain (extrapolation). The hyperparameters for the AE and CAE architectures are listed in Table \ref{tab:table7}.

\begin{table}[h]
    \centering
    \begin{tabular}{c c c}
         \hline
         Hyperparameters & MLP AE & Convolutional CAE\\ [0.5ex] 
         \hline\hline
         Encoder layers & [500, 250] & [8, 32, 32] \\ 
         \hline
         latent dimension ($m$) & 25, 50, 125 & 25,50,125 \\
         \hline
         Activation & relu, swish &relu, swish\\
         \hline
         Loss Function & MSE & MSE \\
         \hline
         Learning rate & $10^{-3},3 \times 10^{-4}$& $10^{-3},3 \times 10^{-4}$\\
         \hline
    \end{tabular}
    \caption{Hyperparameters for the AE and CAE networks}
    \label{tab:table7}
\end{table}

\subsubsection{LSTM}
\begin{figure}[t!]
    \centering
    \begin{subfigure}{\textwidth}
    \begin{subfigure}{0.33 \textwidth}
    \centering
    \includegraphics[width = \textwidth]{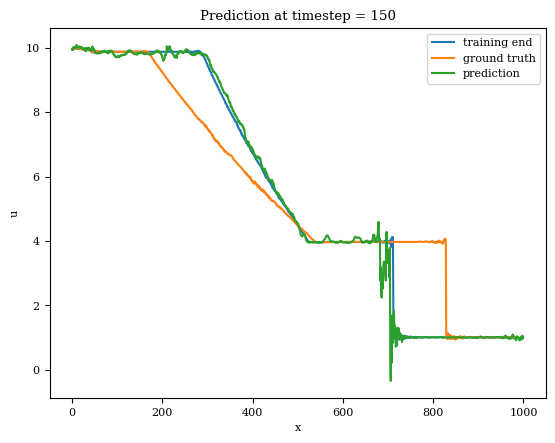}
    \end{subfigure}%
    \begin{subfigure}{0.33 \textwidth}
    \centering
    \includegraphics[width = \textwidth]{CAE_lstm_Stok_predgraphPlot150_epoch2350.png}
    \end{subfigure}%
    \begin{subfigure}{0.33 \textwidth}
    \centering
    \includegraphics[width = \textwidth]{CAE_lstm_Stok_predgraphPlot150_epoch2350.png}
    \end{subfigure}%
    \end{subfigure}%
    \caption{Stoker's problem: Extrapolative auto-regressive predictions of  the LSTM model for time-steps = 310, 360 and 410; the end of training time-step = 260 }
    
    \label{fig:17}
\end{figure}

\begin{figure}[t!]
    \centering
    \centering
    \begin{subfigure}{.5 \textwidth}
    \centering
    \includegraphics[width = \textwidth]{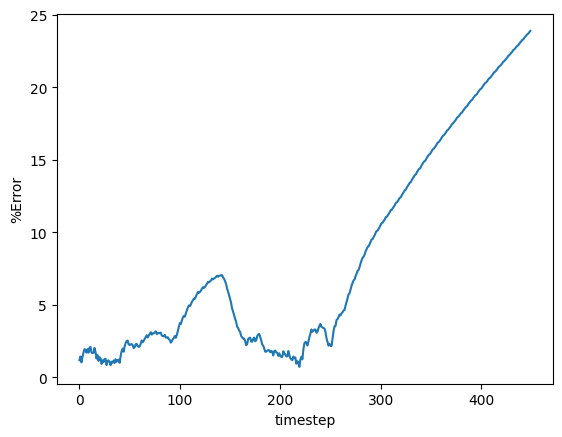}
    \end{subfigure}%
    \caption{Stoker's problem" $L_2$ relative error of the auto-regressive predictions with increasing time for the LSTM model}
    \label{fig:18}
\end{figure}

The LSTM model receives input $Z \,  \epsilon \, \mathbb{R}^{m \times n_t}$ and predicts $z^{i+1}\, \epsilon \, \mathbb{R}^{m} $ (Section \ref{subsubsec:LSTM}). The LSTM model has an architecture similar to that of the Burgers' case, with multiple LSTM layers having a latent dimension as their hidden dimension. The hyperparameters of the LSTM network for the Stoker's problem are summarized in Table \ref{tab:table8}.

\begin{table}[t]
    \centering
    \begin{tabular}{c c c}
         \hline
         Hyperparameters & Values\\ [0.5ex] 
         \hline\hline
         Sequence length ($n_t$) & 5, 10, 20\\
         \hline
         LSTM layers & 1, 2, 3\\ 
         \hline
         hidden/latent dimension ($m$) & 25, 50, 125 \\
         \hline
         Activation & tanh\\
         \hline
         Loss Function & MSE\\
         \hline
         Learning rate & $5 \times 10^{-4}$\\
         \hline
    \end{tabular}
    \caption{Hyperparameters for the LSTM network}
    \label{tab:table8}
\end{table}

The models are trained in batches of size 15, and the loss values for both training and validation converge in 2400 epochs. The model with the least validation loss has a lookback window of size 10 and 2 LSTM layers with hidden dimensions of 125 each. The extrapolation (Figure \ref{fig:17}) and error plots (Figure \ref{fig:18}) obtained from these models show that the LSTM model accurately estimates the solution vectors for time-steps within the training domain ($i <= 250$), but fails outside the training domain, as the relative error reaches 25\% for the 150th time-step post-training.

\subsubsection{TCN model}
The TCN model also receives a sequence $Z \, \epsilon \, \mathbb{R}^{m \times n_t}$ and forecasts $z^{i+1}\, \epsilon \, \mathbb{R}^{m} $, with the dilated convolutions operating on the temporal axis, and the latent dimension passed as a channel. The model contains three temporal blocks (Section \ref{subsubsec:TCN}), with the same kernels and channels in each, and dilations that increase in size by a factor of two in subsequent blocks. The hyperparameters for the TCN network are listed in Table \ref{tab:table9}.

\begin{table}[h]
    \centering
    \begin{tabular}{c c c}
         \hline
         Hyperparameters & Values\\ [0.5ex] 
         \hline\hline
         Sequence length ($n_t$) & 5, 10, 20\\
         \hline
         TCN block channels & [100, 100, 100], [200, 200, 200]\\ 
         \hline
         latent dimension ($m$) & 25,50,125 \\
         \hline
         Kernel Size(k) & 3, 5, 7, 9 \\
         \hline
         Activation & tanh\\
         \hline
         Loss Function & MSE\\
         \hline
         Learning rate & $3 \times 10^{-4}$\\
         \hline
    \end{tabular}
    \caption{Hyperparameters for the TCN network}
    \label{tab:table9}
\end{table}
\begin{figure}[t!]
    \centering
    \begin{subfigure}{\textwidth}
    \begin{subfigure}{0.33 \textwidth}
    \centering
    \includegraphics[width = \textwidth]{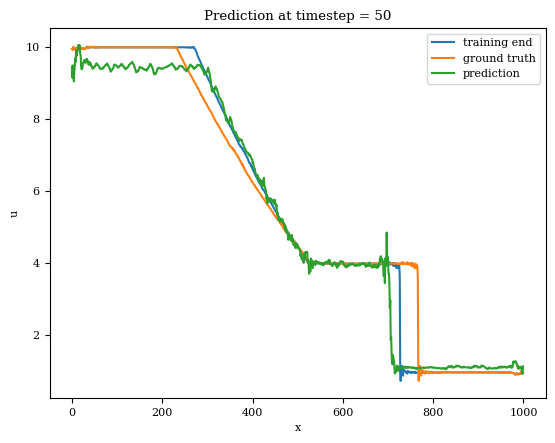}
    \end{subfigure}%
    \begin{subfigure}{0.33 \textwidth}
    \centering
    \includegraphics[width = \textwidth]{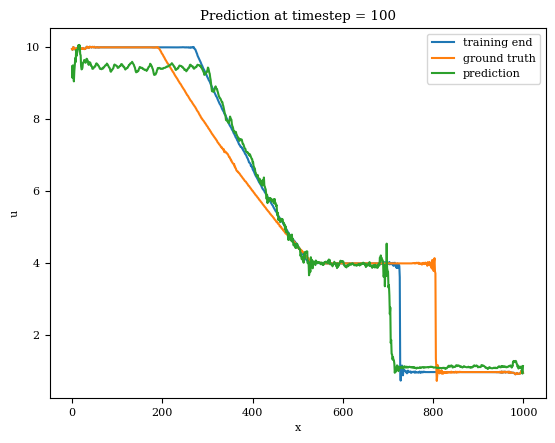}
    \end{subfigure}%
    \begin{subfigure}{0.33 \textwidth}
    \centering
    \includegraphics[width = \textwidth]{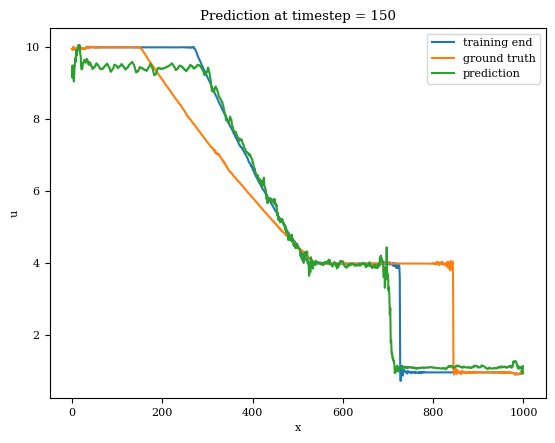}
    \end{subfigure}%
    \end{subfigure}%
    
    \caption{Stoker's problem: Extrapolative auto-regressive predictions using the TCN model  for time-steps = 320, 370 and 420. The training end is at time-step = 270)}
    
    \label{fig:19}
\end{figure}

\begin{figure}[t!]
    \centering
    \centering
    \begin{subfigure}{.5 \textwidth}
    \centering
    \includegraphics[width = \textwidth]{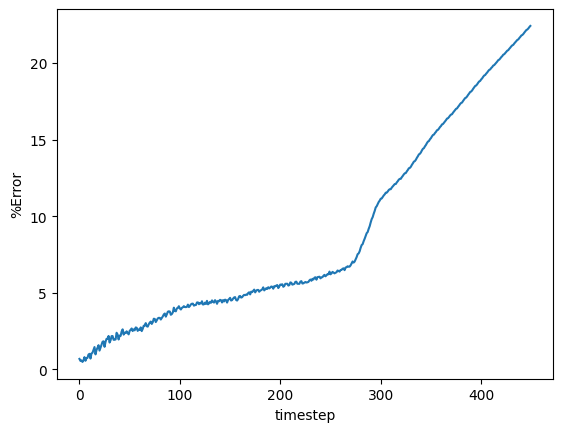}
    \end{subfigure}%
    \caption{Stoker's problem: $L_2$  relative error of the auto-regressive predictions with increasing time for the TCN model }
    \label{fig:20}
\end{figure}
Model training and validation loss reaches convergence by 1000 epochs, when training and validation is performed in batches of size 15. The best model (with the least validation loss) accommodates a  sequence of vectors with a lookback window of size 20, and 125 latent modes. Each temporal block contains kernels of size 3 and has 200 channels. The extrapolation (Figure \ref{fig:19}) and error plots (Figure \ref{fig:20}) obtained from these models indicate that similar to the LSTM, the TCN also predicts the solution vectors with acceptable accuracy for time-steps within the training domain ($i <= 250$), but the solution stops propagating further in the extrapolative domain, and so  the error increases to 23\%.

\subsubsection{CNN model}

\begin{figure}[b!]
    \centering
    \begin{subfigure}{\textwidth}
    \begin{subfigure}{0.33 \textwidth}
    \centering
    \includegraphics[width = \textwidth]{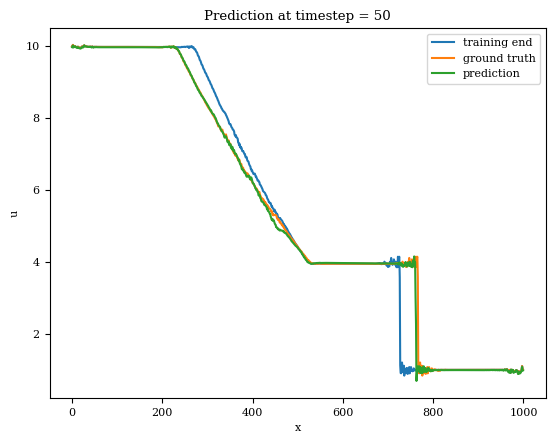}
    \end{subfigure}%
    \begin{subfigure}{0.33 \textwidth}
    \centering
    \includegraphics[width = \textwidth]{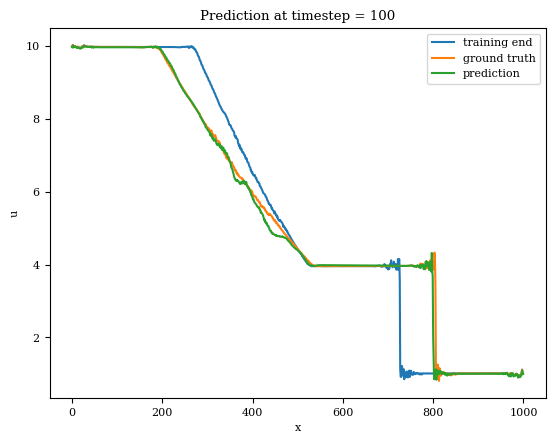}
    \end{subfigure}%
    \begin{subfigure}{0.33 \textwidth}
    \centering
    \includegraphics[width = \textwidth]{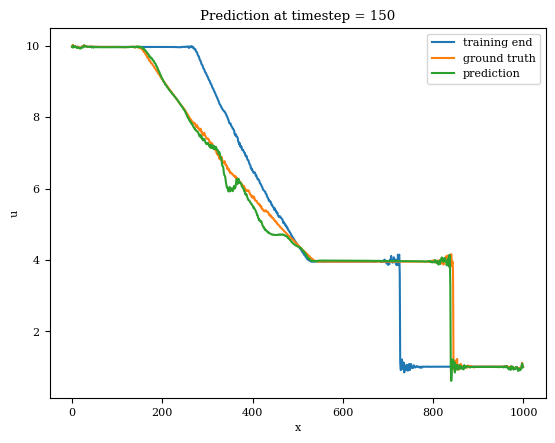}
    \end{subfigure}%
    \end{subfigure}%
    
    \caption{Stoker's problem: Extrapolative auto-regressive predictions using the CNN model for time-steps = 320, 370 and 420; the end of training is at time-step = 270 }
    
    \label{fig:21}
\end{figure}

\begin{figure}[t!]
    \centering
    \centering
    \begin{subfigure}{.5 \textwidth}
    \centering
    \includegraphics[width = \textwidth]{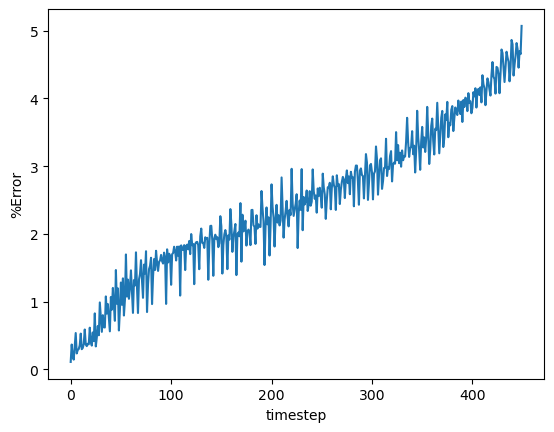}
    \end{subfigure}%
    \caption{Stoker's problem, $L_2$  relative error of the auto-regressive predictions with increasing time for the CNN model}
    \label{fig:22}
\end{figure}

The proposed CNN model takes the transposed vector sequence $Z^T \, \epsilon \, \mathbb{R}^{n_t \times m}$ to produce $z^{i+1}\, \epsilon \, \mathbb{R}^{m} $, with 1D convolutions operating on the spatial axis or latent dimension, and the $n_t$ latent vectors present on separate channels. The CNN model consists of three residual blocks (Section \ref{subsubsec:CNN}), each with the same kernel size and number of channels. The hyperparameters for the CNN model are summarized in Table \ref{tab:table10}:

\begin{table}[h]
    \centering
    \begin{tabular}{c c c}
         \hline
         Hyperparameters & Values\\ [0.5ex] 
         \hline\hline
         Sequence length ($n_t$) & 5, 10, 20\\
         \hline
         TCN block channels & [50, 50, 50], [100, 100, 100], [200, 200, 200]\\ 
         \hline
         latent dimension ($m$) & 25, 50, 125 \\
         \hline
         Kernel Size(k) & 3, 5, 7, 9 \\
         \hline
         Activation & tanh\\
         \hline
         Loss Function & MSE\\
         \hline
         Learning rate & $3 \times 10^{-4}$\\
         \hline
    \end{tabular}
    \caption{Hyperparameters for the CNN model}
    \label{tab:table10}
\end{table}

The training and validation loss converges at around 1000 epochs when batches of size 16 are used. The model with the highest accuracy (least validation loss and RA) has a lookback window of 20 steps, with 125 nodes in latent vectors at every step. Each of the three blocks possess a kernel of size 3 and 100 channels in the 1D convolution layers. The extrapolation (Figure  \ref{fig:21}) and error plots (Figure \ref{fig:22}) indicate that the CNN model is capable of modelling the latent dynamics, since accurate forecasts are produced for time-steps beyond the training domain. The error values increase over time, reaching 5\%, which is remarkably lower than that of earlier models.

\section{Uncertainty Quantification using Deep Ensembles}

\begin{figure}[t!]
    \centering
    \begin{subfigure}{\textwidth}
    \begin{subfigure}{0.33 \textwidth}
    \centering
    \includegraphics[width = \textwidth]{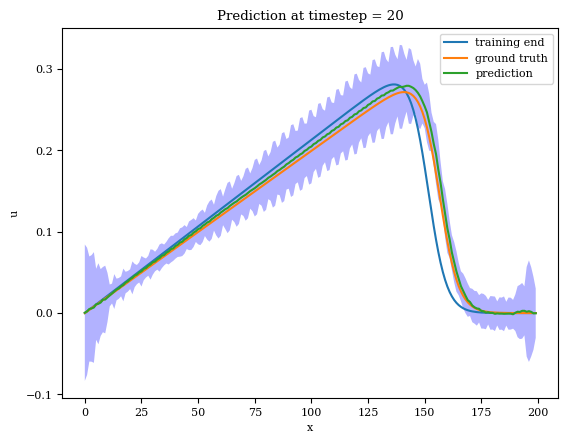}
    \end{subfigure}%
    \begin{subfigure}{0.33 \textwidth}
    \centering
    \includegraphics[width = \textwidth]{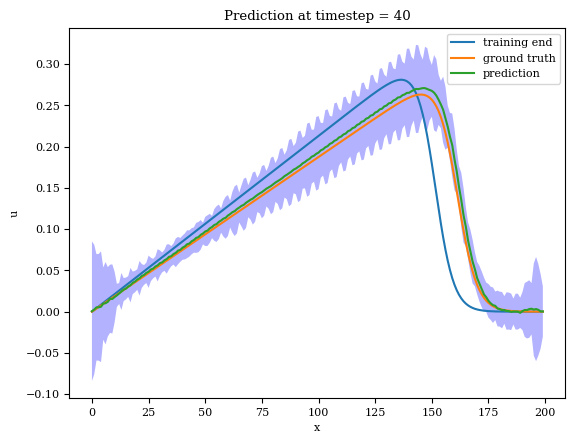}
    \end{subfigure}%
    \begin{subfigure}{0.33 \textwidth}
    \centering
    \includegraphics[ width= \textwidth]{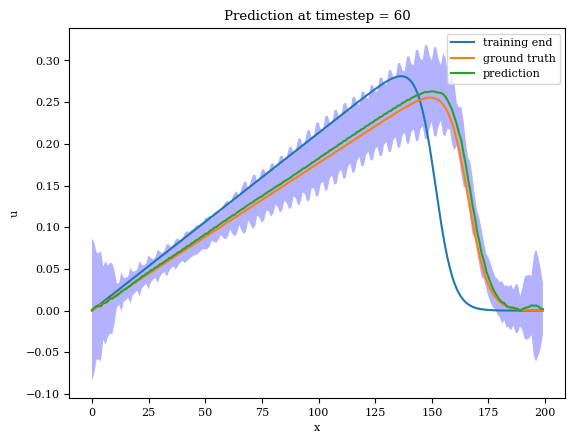}
    \end{subfigure}%
    \caption{Re = 300}
    \end{subfigure}%
    \\
    \begin{subfigure}{\textwidth}
    \begin{subfigure}{0.33 \textwidth}
    \centering
    \includegraphics[width = \textwidth]{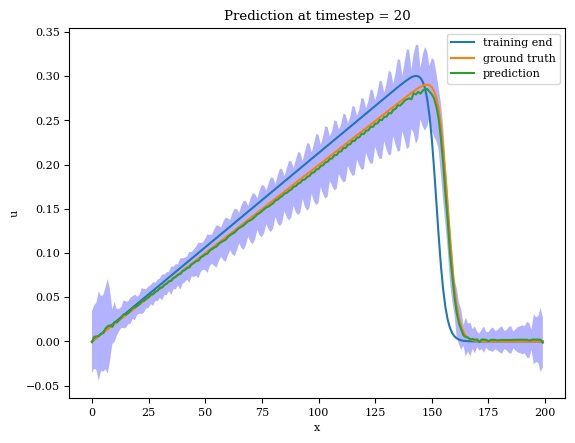}
    \end{subfigure}%
    \begin{subfigure}{0.33 \textwidth}
    \centering
    \includegraphics[width = \textwidth]{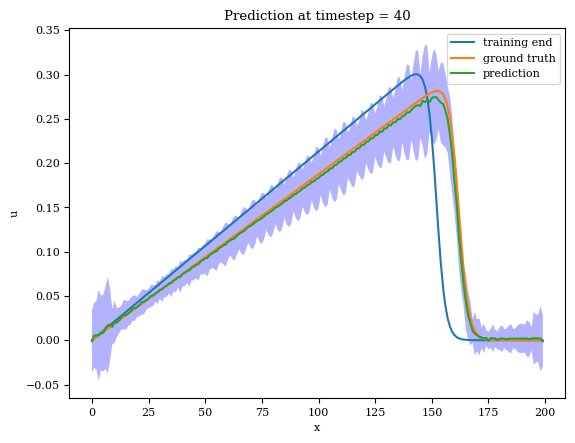}
    \end{subfigure}%
    \begin{subfigure}{0.33 \textwidth}
    \centering
    \includegraphics[width = \textwidth]{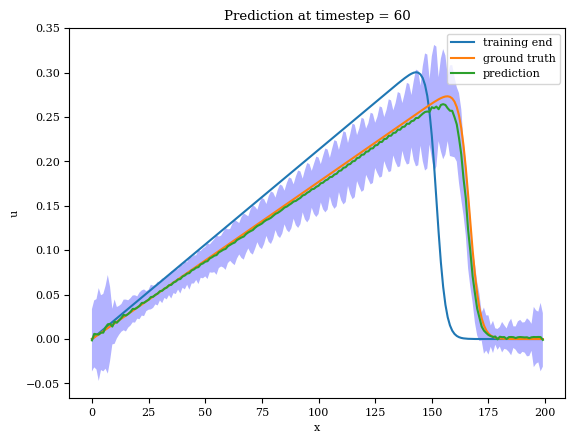}
    \end{subfigure}%
    \caption{Re = 600}
    \end{subfigure}%
    \caption{Burgers' problem: Extrapolative auto-regressive predictions by the deep ensemble CAE-CNN model for time-steps = 180, 200 and 220 and for Re= 300 (a) and Re= 600 (b); the end of  training is at time-step=160}
    
    \label{fig:23}
\end{figure}

\begin{figure}[t!]
    \centering
    \centering
    \begin{subfigure}{.5 \textwidth}
    \centering
    \includegraphics[width = \textwidth]{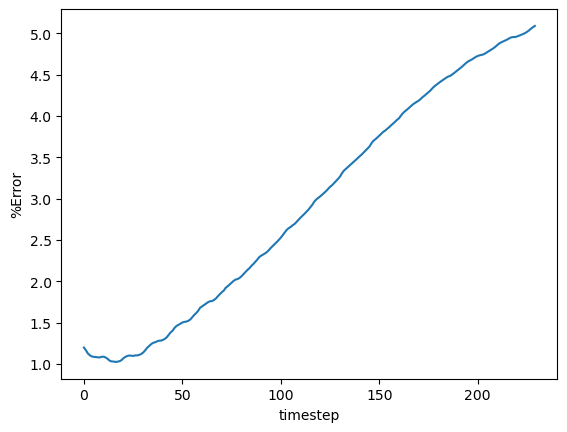}
    \caption{Re = 300}
    \end{subfigure}%
    \begin{subfigure}{.5 \textwidth}
    \centering
    \includegraphics[width = \textwidth]{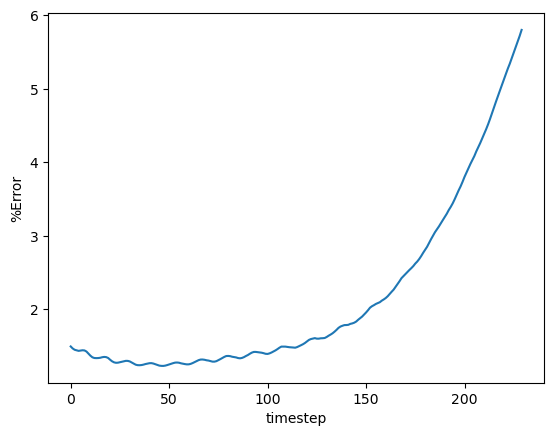}
    \caption{Re = 600}
    \end{subfigure}
    \caption{Burgers' problem: $L_2$  relative error of the auto-regressive predictions with increasing time for Re = 300 (a) and Re = 600 (b) for a deep ensemble CAE-CNN model}
    \label{fig:24}
\end{figure}

The training-testing methodology described previously{\ref{subsec:ensemble}} for the evaluation of variance in the predicted output is adopted for the most stable (i.e., similar results for discrete initialization) and most accurate  (i.e., least relative error on extrapolation) forecasting model,  (i.e. the CNN future step predictor with a CAE autoencoder). After the model is trained to minimize the NLL, the mean and variance for the latent vector,  $\mu^z_*$ and ${\sigma^z_*}^2$ respectively, are calculated using predictions from $M = 10$ models. The mean and variance, $\mu^v_*$ and ${\sigma^v_*}^2$ for predicted vectors $[v^{n_{t}+1}, v^{n_{t}+2}, …, v^{T}]$ are then calculated using $2m+1$ sigma points sampled (with $k = 0.2$) on $\hat{z}$ and decompressed in accordance with Equation \ref{eq14} and \ref{eq15}.

\begin{figure}[t!]
    \centering
    \begin{subfigure}{0.33 \textwidth}
    \centering
    \includegraphics[width = \textwidth]{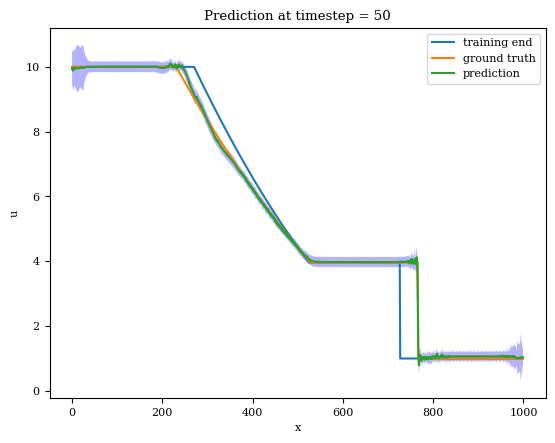}
    \end{subfigure}%
    \begin{subfigure}{0.33 \textwidth}
    \centering
    \includegraphics[width = \textwidth]{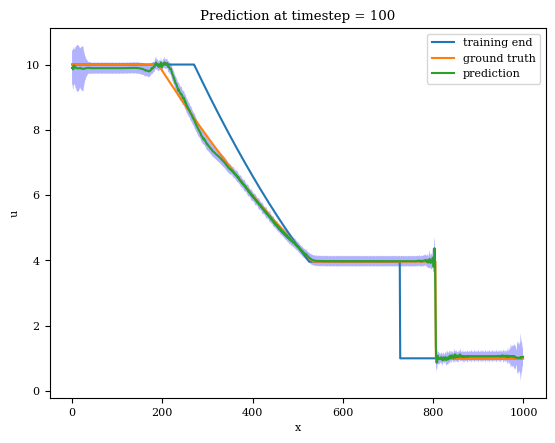}
    \end{subfigure}%
    \begin{subfigure}{0.33 \textwidth}
    \centering
    \includegraphics[width = \textwidth]{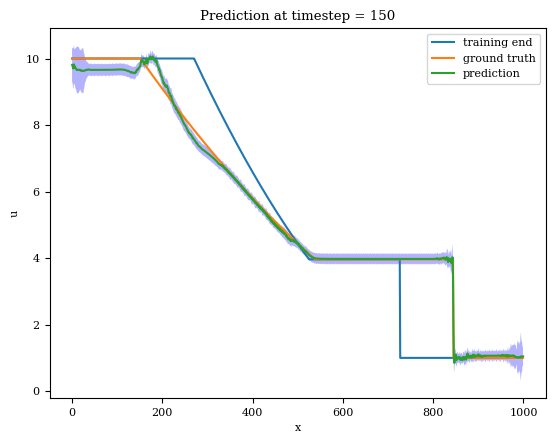}
    \end{subfigure}%
    \caption{Stoker's problem: Extrapolative auto-regressive predictions by a deep ensemble CAE-CNN model for time-steps = 320, 370 and 420;  the end of training is at  time-step = 270}
    
    \label{fig:25}
\end{figure}

The extrapolation results obtained by the ensemble are demonstrated in Figure \ref{fig:23} for the 1D Burgers' test case with Re = 300 and 600, and in Figure \ref{fig:25} for the 1D Stoker's test case.

\begin{figure}[t!]
    \centering
    \centering
    \begin{subfigure}{.5 \textwidth}
    \centering
    \includegraphics[width = \textwidth]{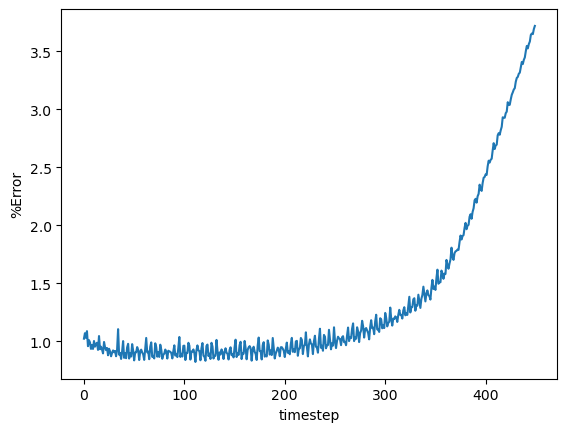}
    \end{subfigure}%
    \caption{Stoker's problem: $L_2$  relative error of the auto-regressive predictions with increasing time for the deep ensemble CAE-CNN model}
    \label{fig:26}
\end{figure}
The predicted mean for all time-steps outside the training domain overlaps with the actual value, and the error reaches a maximum of 6\% (Figure \ref{fig:24}) for the Burgers' and 3.5\% (Figure \ref{fig:26}) for the Stoker's problem. The variance is maximum at the spatial terminal ends of the predicted vector solutions, as  they lack neighbouring nodes on one side, which are essential for  information extraction using 1D convolution filters for accurate future step estimations. The variance remains almost constant throughout the prediction time, indicating the robustness of the proposed model.

\section{Conclusion}

This study proposes a Convolutional Autoencoder(CAE) model for compression and a CNN future step predictor for forecasting vector solutions for subsequent time-steps to input vector sequences. The approximation accuracy and time extrapolation capabilities of the model are evaluated using two advection-dominated flow problems, a 1D Burgers' equation and a 1D Stoker's equation, which are characterized by sharp gradients and discontinuities, respectively. The models built especially for time-series forecasts, LSTM and TCN models with propagation over time, produce acceptable results within the training domain, but the solution stops changing during extrapolation. However, when the dilated convolutions propagate on the spatial axis, the models produce good predictions for extrapolation as well. The proposed CNN model for forecasting has an architecture (residual blocks with 1D convolutions propagating along space) similar to that of the TCN model, but without causal padding or dilation. These have been eliminated, as the increasing receptive field has no significance during convolution in space, and  causal padding degrades the results by causing  the information extracted from neighbouring cells to be shifted or scraped. The CNN model is capable of producing highly accurate predictions for both test cases, with less than 5\% relative $L_2$ error in the extrapolation domain ($\approx$ 60\% of the training time domain). However, the CNN model only produces accurate forecasts if it receives compressed latent vectors from the CAE model, since it preserves the local spatial information better than the MLP encoder during compression. Uncertainty quantification for the variance-informed CAE-CNN model is performed using deep ensembles, which  produce accurate predictions, with low variance for long-term extrapolates as well (errors less than 6\%). In addition, since the CNN model uses convolutions (like TCN \cite{BaiTCN2018}), the training and evaluation can be done in parallel for long input sequences, in contrast to the LSTM. Thus, a fast, accurate and robust framework is provided for order reduction. The model architecture is flexible and can be extended for two- and three-dimensional spaces as well, by increasing the dimension of the convolutional filters. 

It would  be interesting to test if the CAE-CNN architecture can adapt to problems where the solutions/data for the model are obtained using unstructured meshes. Future work could  also focus on adapting the architecture to resolve real-life engineering problems. 

\section*{Acknowledgments}
This research was supported by the Natural Sciences and Engineering Research Council of Canada and MITACS; the financial support is gratefully acknowledged. The computing resources were provided by the Digital Research Alliance of Canada, whose support is highly appreciated.

\bibliographystyle{unsrt}  
\bibliography{references}  

\appendix

\section{Heat Map Plots for Burgers' and Stoker's problems}

\label{sec:heat map appendix burgers and stokers}

\begin{table}[h!]
    \centering
    \begin{tabular}{c c c}
         \hline
         Forecasting Models & MLP AE & Convolutional CAE\\ [0.5ex] 
         \hline
         \multirow{2}{*}{Ground Truth} & \includegraphics[width=0.15\linewidth]{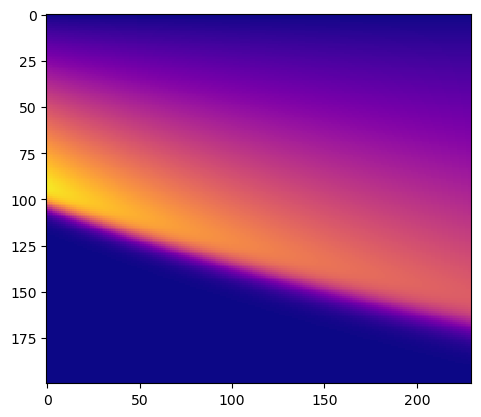}
         \includegraphics[width=0.025\linewidth]{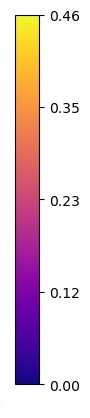}
         & \includegraphics[width=0.15\linewidth]{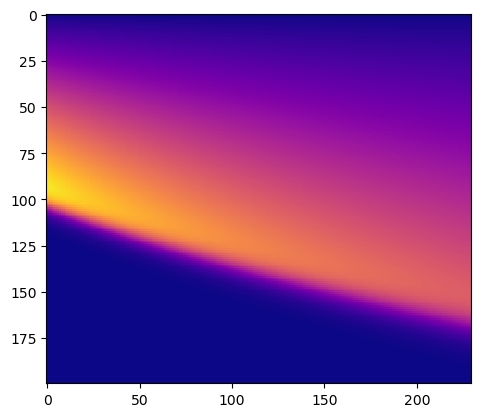}
         \includegraphics[width=0.025\linewidth]{Burger300_bar.jpeg}\\
         & \includegraphics[width=0.15\linewidth]{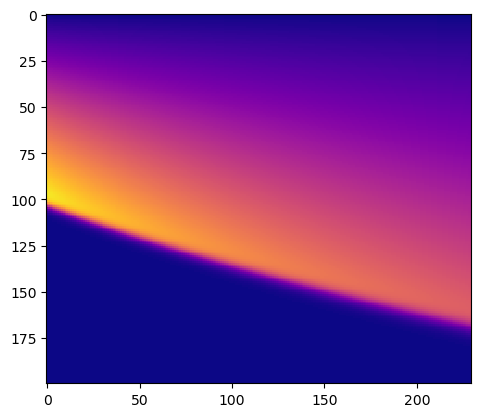}
         \includegraphics[width=0.025\linewidth]{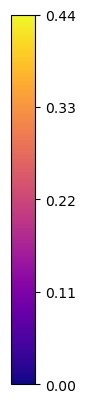}
         & \includegraphics[width=0.15\linewidth]{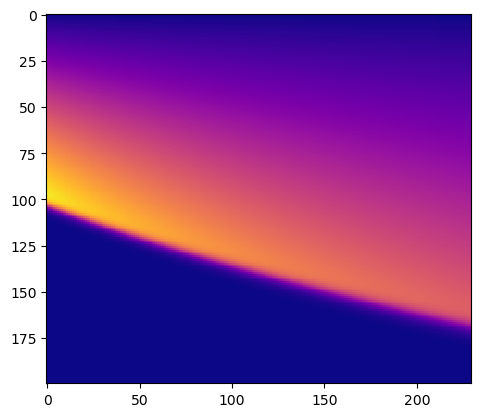}
         \includegraphics[width=0.025\linewidth]{Burger600_bar.jpeg}\\
         \hline
         \multirow{2}{*}{LSTM} & \includegraphics[width=0.15\linewidth]{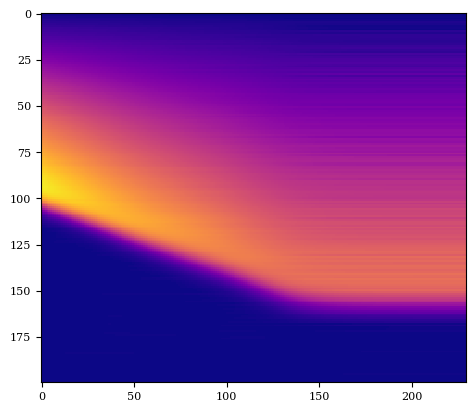} & \includegraphics[width=0.15\linewidth]{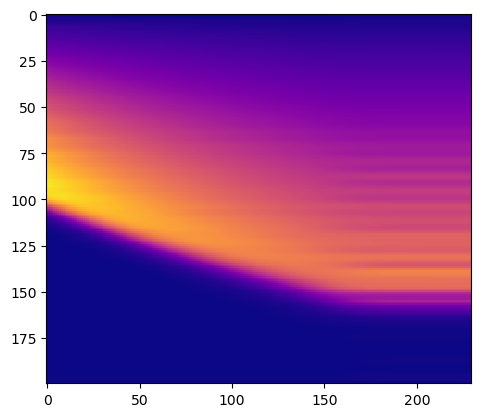}\\
         & \includegraphics[width=0.15\linewidth]{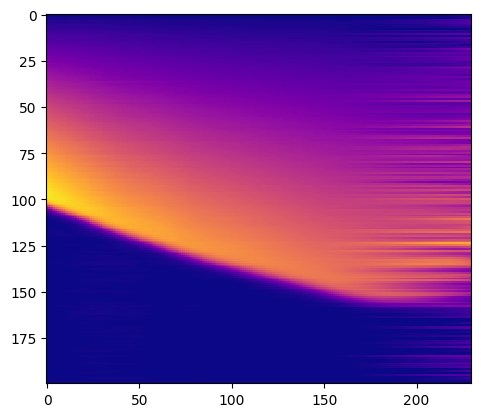} & \includegraphics[width=0.15\linewidth]{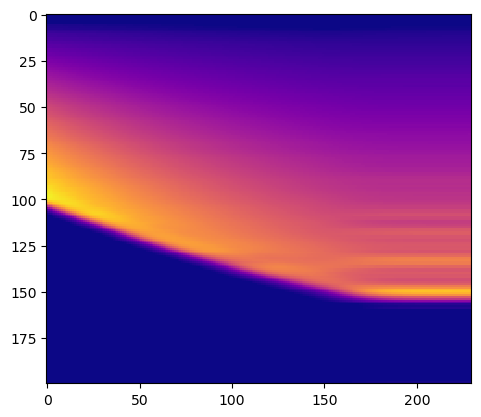}\\
         \hline
         \multirow{2}{*}{TCN} & \includegraphics[width=0.15\linewidth]{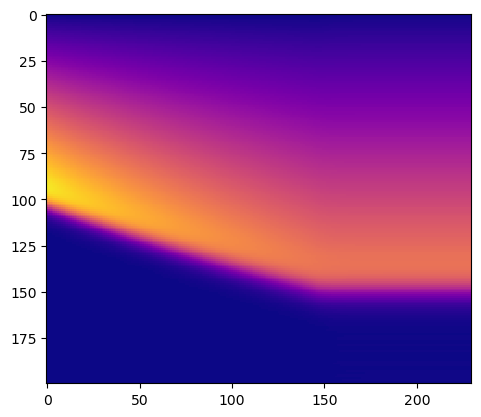} & \includegraphics[width=0.15\linewidth]{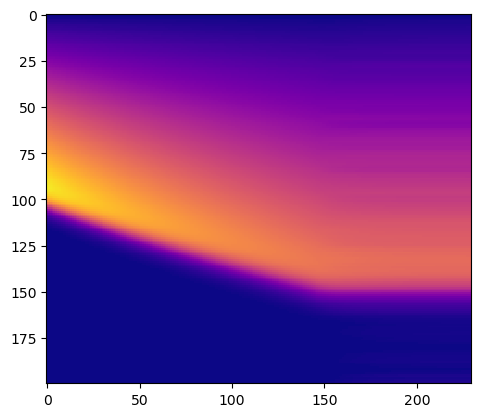}\\
         & \includegraphics[width=0.15\linewidth]{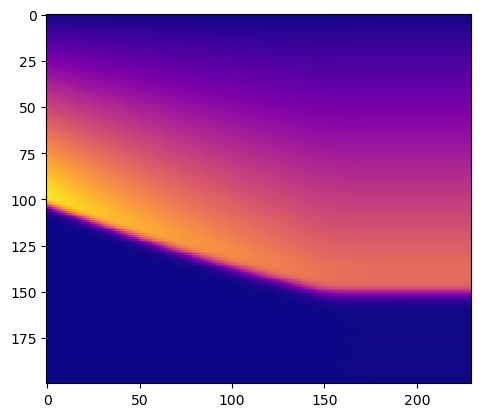} & \includegraphics[width=0.15\linewidth]{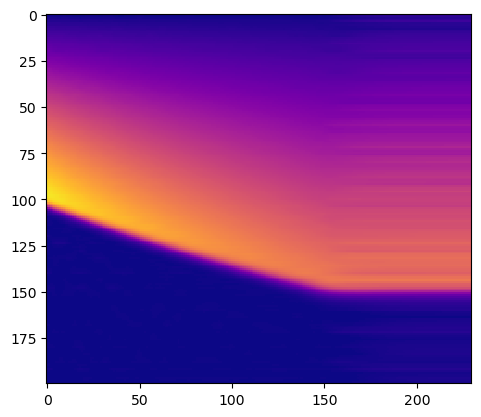}\\
         \hline
         \multirow{2}{*}{CNN} & \includegraphics[width=0.15\linewidth]{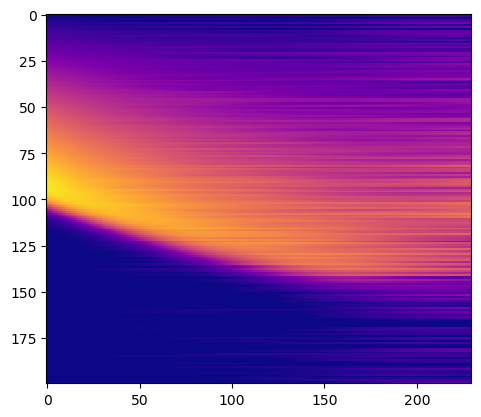} & \includegraphics[width=0.15\linewidth]{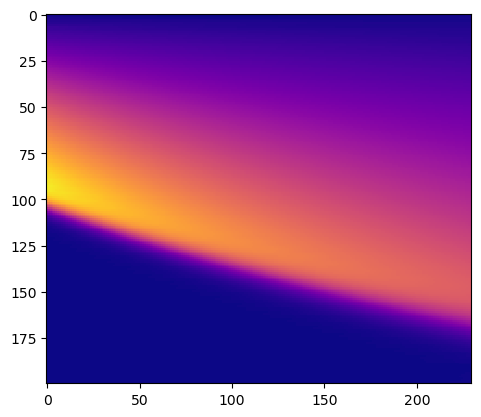}\\
         & \includegraphics[width=0.15\linewidth]{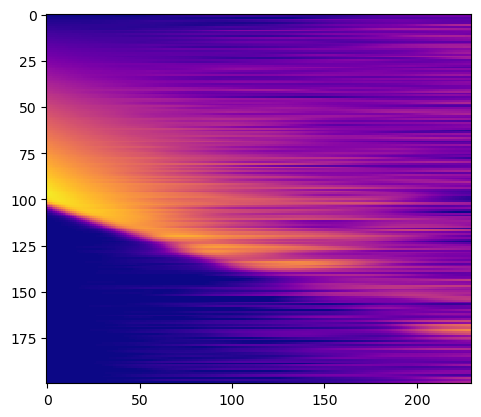} & \includegraphics[width=0.15\linewidth]{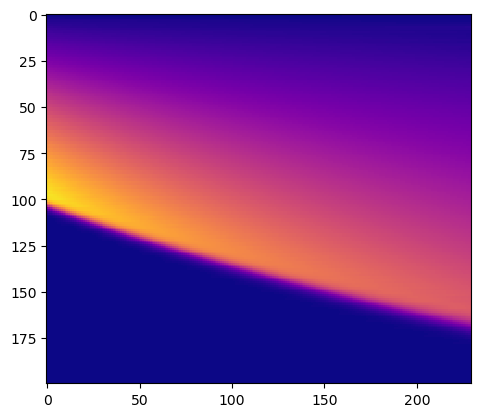}\\
         \hline
    \end{tabular}
    \caption{Burgers' problem, auto-regressive forecasts for Re= 300 (top) and Re= 600 (bottom) from the forecasting models when latent vectors are obtained by AE and CAE}
    \label{tab:A.1}
\end{table}

\begin{table}[h!]
    \centering
    \begin{tabular}{c c c}
         \hline
         Forecasting Models & MLP AE & Convolutional CAE\\ [0.5ex] 
         \hline
         \centering Ground Truth & \includegraphics[width=0.15\linewidth]{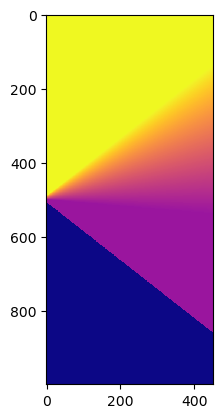} \includegraphics[width=0.06\linewidth]{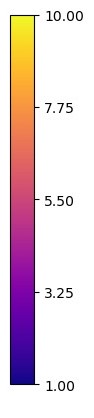}& \includegraphics[width=0.15\linewidth]{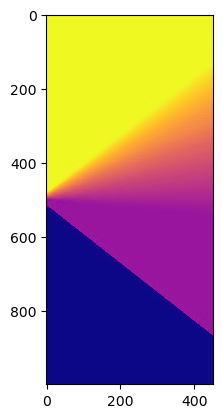}
         \includegraphics[width=0.06\linewidth]{Stoker_bar.jpeg}
         \\
         \hline
         LSTM & 
         \includegraphics[width=0.15\linewidth]{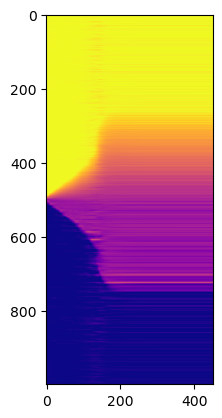} & \includegraphics[width=0.15\linewidth]{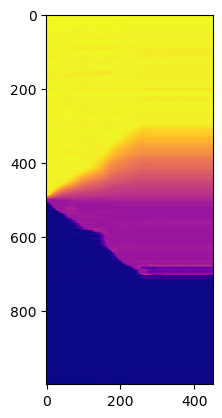}\\
         \hline
         TCN & 
         \includegraphics[width=0.15\linewidth]{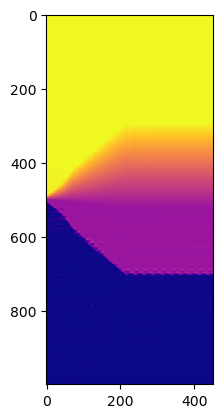} & \includegraphics[width=0.15\linewidth]{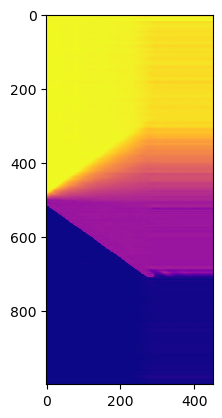}\\
         \hline
         CNN & 
         \includegraphics[width=0.15\linewidth]{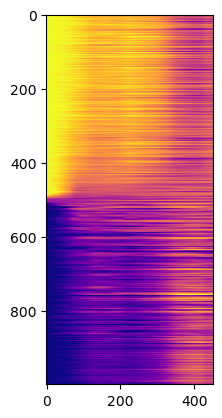} & \includegraphics[width=0.15\linewidth]{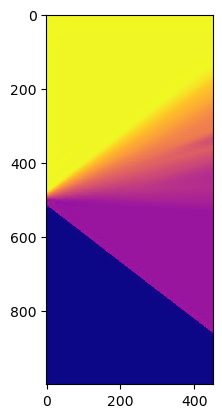}\\
         \hline
    \end{tabular}
    \caption{Auto-regressive forecasts for the Stoker's problem from the forecasting models when latent vectors are obtained by AE and CAE}
    \label{tab:A.2}
\end{table}

\end{document}